\documentclass[runningheads]{llncs}
\usepackage{graphicx}

\usepackage{tikz}
\usepackage{comment}
\usepackage{amsmath,amssymb} 
\usepackage{color}
\usepackage{multirow}
\usepackage{tabularx}

\usepackage[accsupp]{axessibility}  

\begin{document}
\pagestyle{headings}
\mainmatter
\def\ECCVSubNumber{5470}  

\title{3D-C2FT: Coarse-to-fine Transformer for Multi-view 3D Reconstruction} 

\titlerunning{3D-C2FT: Coarse-to-fine Transformer for Multi-view 3D Reconstruction}
%
\author{Leslie Ching Ow Tiong\inst{1} \and
Dick Sigmund\inst{2} \and
Andrew Beng Jin Teoh\inst{3}}
\authorrunning{L. Tiong et al.}
%
\institute{Computational Science Research Center, Korea Institute of Science and Technology, 5, Hwarang-ro 14-gil, Seongbuk-gu, Seoul 02792, Republic of Korea\\
\email{tiongleslie@kist.re.kr} \and
Aidot Inc., 128, Beobwon-ro, Songpa-gu, Seoul 05854, Republic of Korea\\
\email{dsigmund@aidot.ai} \and
School of Electrical and Electronic Engineering, Yonsei University, Seoul 120-749, Republic of Korea\\
\email{bjteoh@yonsei.ac.kr}}
\maketitle

\begin{abstract}
Recently, the transformer model has been successfully employed for the multi-view 3D reconstruction problem. However, challenges remain on designing an attention mechanism to explore the multi-view features and exploit their relations for reinforcing the encoding-decoding modules. This paper proposes a new model, namely 3D coarse-to-fine transformer (3D-C2FT), by introducing a novel coarse-to-fine (C2F) attention mechanism for encoding multi-view features and rectifying defective 3D objects. C2F attention mechanism enables the model to learn multi-view information flow and synthesize 3D surface correction in a coarse to fine-grained manner. The proposed model is evaluated by ShapeNet and Multi-view Real-life datasets. Experimental results show that 3D-C2FT achieves notable results and outperforms several competing models on these datasets.
\end{abstract}

\section{Introduction}
\label{sec::sec1}

Multi-view 3D reconstruction infers geometry and structure of voxelized 3D objects from single or multi-view images. This task plays a crucial role in robot perception \cite{Burchfiel2017}, \cite{Shi2017}, historical artifact \cite{Kargas2019}, \cite{Nabil2014}, dentistry \cite{Pavaloiu2014}, \cite{Roointan2019}, etc. Predicaments in estimating 3D structure of an object include solving an ill-posed inverse problem of 2D images, violation of overlapping views, and unconstrained environment conditions \cite{Han2021}. Here, the ill-posed inverse problem refers to the voxelized 3D objects estimation problem when disordered or limited sources of the inputs are given as inputs \cite{Groen2019}. Accordingly, multi-view 3D object reconstruction remains an active topic in computer vision. 

Classical approaches to tackle multi-view 3D reconstruction are to introduce feature extraction techniques to map different reconstruction views \cite{Gao2014}, \cite{Silveira2008}, \cite{Tron2014}, \cite{Wilson2014}. However, these approaches hardly reconstruct a complete 3D object when a small number of images, e.g., 1--4, are presented. With the advances of deep learning (DL), most attempts utilize an encoder-decoder network architecture for feature extraction, fusion, and reconstruction. Existing DL models usually rely on the convolutional neural networks (CNN) or recurrent neural networks (RNN) to fuse multiple deep features encoded from 2D multi-view images \cite{Choy2016}, \cite{Tatarchenko2017}. However, the CNN encoder processes each view image independently, and thus, the relations in different views are barely utilized. This leads to difficulty designing a fusion method that can traverse the relationship between views. Although the RNNs can rectify the fusion problem, the model suffers a long processing time \cite{Choy2016}.

Recently, transformers \cite{Vaswani2017} have gained exponential attention and proved to be a huge success in vision-related problems \cite{Dosovitskiy2021}, \cite{Park2022}. In the multi-view 3D reconstruction problem, \cite{Wang2021} and \cite{Yagubbayli2021} propose a unified multi-view encoding, fusion, and decoding framework via a transformer to aggregate features among the patch embeddings that can explore the profound relation between multi-view images and perform the decent reconstruction. However, both studies only consider single-scaled multi-head self-attention (MSA) mechanism to reinforce the view-encoding. Unfortunately, the native single-scaled MSA is not designed to explore the correlation of relevant features between the subsequent layers for object reconstruction.

Considering the limitations above, we propose a new transformer model, namely 3D coarse-to-fine transformer (3D-C2FT). The 3D-C2FT introduces a novel coarse-to-fine (C2F) attention mechanism for encoded features aggregation and decoded object refinement in a multi-scale manner. On top of exploring multi-view images relationship via MSA in the native transformer, the C2F attention mechanism enables aggregating coarse (global) to fine (local) features, which is favored to learn comprehensive and discriminative encoded features. Then, a concatenation operation is used to aggregate multiple sequential C2F features derived from each C2F attention block. In addition, a C2F refiner is devised to rectify the defective decoded 3D objects. Specifically, the refiner leverages a C2F cube attention mechanism to focus on the voxelized object's attention blocks that benefit coarse-to-fine correction of the 3D surface. In general, C2F attention utilizes the coarse and fine-grained features to be paired and promotes the information flow that helps the 3D objects reconstruction task.

Besides that, we compile a new dataset called Multi-view Real-life dataset for evaluation. However, unlike existing dataset such as ShapeNet \cite{Wu2015} that is assembled from synthetic data, our dataset is composed of real-life single or multi-view images taken from internet without ground truth. We propose this dataset to support real-life study for multi-view 3D reconstruction evaluation. The contributions of this paper are summarized as follows:
\begin{itemize} 
\item A novel multi-scale C2F attention mechanism is outlined for the multi-view 3D reconstruction problem. The proposed C2F attention mechanism learns the correlation of the features in a sequential global to the local manner, on top of exploring multi-view images relation via MSA. Accordingly, coarse-grained features draw attention towards the global object structure, and the fine-grained feature pays attention to each local part of the 3D object. 
\item A novel transformer model, namely 3D-C2FT, for multi-scale multi-view images encoding, features fusion, and coarse-to-fine decoded objects refinement is proposed.
\item Extensive experiments demonstrate better reconstruction results on standard benchmark ShapeNet dataset and challenging real-life dataset than several competing models, even under stringent constraints such as occlusion.
\end{itemize}

This paper is organized as follows: Section \ref{sec::sec2} reviews the related works of the DL-based multi-view 3D reconstruction. Then, Section \ref{sec::sec3} presents our proposed method. Next, the experimental setup, results, and ablation study are presented in Section \ref{sec::sec4}. Finally, the conclusions are summarized in the last section.

\section{Related Works}
\label{sec::sec2}

This section reviews several relevant works of DL-based multi-view 3D reconstruction.  We refer the readers to a more comprehensive survey on this subject \cite{Han2021}.

\subsection{CNN and RNN-based Models}
Early works \cite{Choy2016} and \cite{Kar2017} utilize modified RNN to fuse information from multi-view images and reconstruct the 3D objects. However, both works have their limitations; for instance, when given the same set of images but in different orders, the model fails to reconstruct the 3D objects consistently. This is due to both models being permutation-variant and relying on the ordered sequence of the input images for feature extraction.

\cite{Xie2019}, \cite{Xie2020} and \cite{Yang2020} propose a CNN-based model to address randomly ordered and long sequence forgetting issues of the RNNs. However, these works follow the divide-and-conquer principle by introducing a CNN-based single-view encoder, a single-view decoder, and a fusion model, which work independently. Therefore, the encoder and decoder hardly utilize the relations between different views. Instead, the network relies on the fusion model to integrate the arbitrary ordered multi-view features for reconstruction. Results suggest that it is challenging to design a robust fusion method with a CNN-based approach. Furthermore, these works suffer from the model scaling problem while preserving permutation-invariant capability. For instance, if the input views exceed a specific number, such as 10--16 views, the reconstruction performance level-off, implying the hardness of learning complementary knowledge from a large set of independent inputs.

\subsection{Transformer-based Models} 
Recently, \cite{Wang2021} and \cite{Yagubbayli2021} put forward the transformer-based models that perceive multi-view 3D reconstruction problems as a sequence-to-sequence prediction problem that permits fully exploiting information from 2D images with different views. \cite{Wang2021} leverages native multi-head self-attention (MSA) mechanism for feature encoding, fusing, and decoding to generate a 3D volume for each query token. However, this model only relies on the MSA to explore the relation of multi-view images by fostering different representations of each view. Hence, it falls short in analyzing the low to high-level interactions within multi-view images that signify the global structure and local components of the 3D objects.

\cite{Yagubbayli2021} proposes another transformer model simultaneously along with \cite{Wang2021} known as Legoformer. LegoFormer adopts an encoder with pre-norm residual connections \cite{Nguyen2019} and a non-autoregressive decoder that takes advantage of the decomposition factors without explicit supervision. Although this approach is more effective, the model only focuses on learning 3D objects reconstruction parts by parts. Such a strategy is deemed local-to-local attention, which does not benefit low-level interaction to support information flow across local parts of 3D objects.

Considering the limitations of the previous transformer-based models, we are motivated to design a new attention mechanism by introducing a C2F patch attention mechanism in the encoder to extract multi-scale features of the multi-view images. Furthermore, we also put forward a C2F cube attention mechanism in the refiner to rectify the reconstructed object surface in a coarse to fine-grained manner.

\section{3D Coarse-to-fine Transformer}
\label{sec::sec3}
\begin{figure}[!t]
\centering
\includegraphics[width=0.75\textwidth]{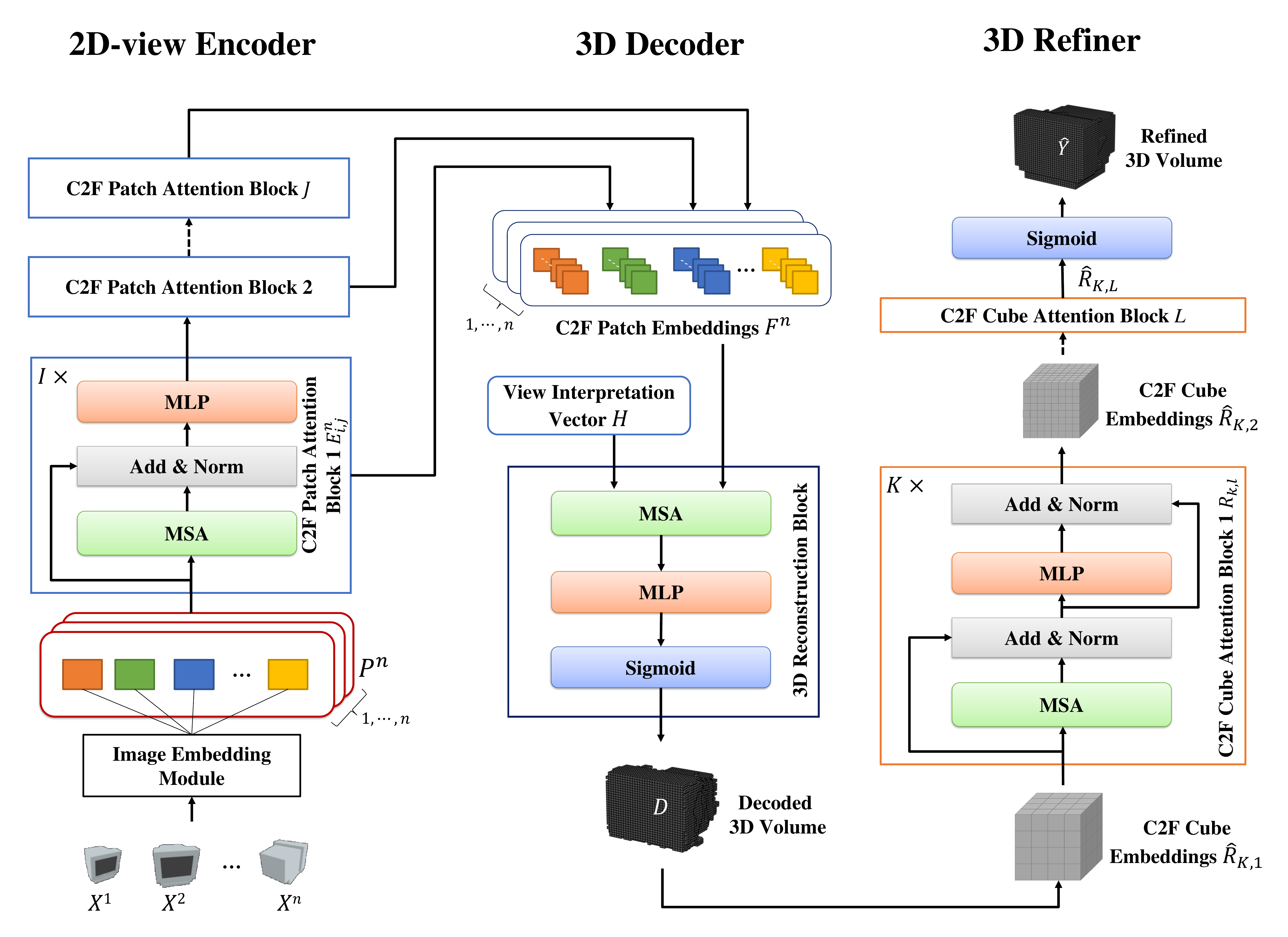}
\caption{The architecture of 3D-C2FT. The network comprises an image embedding module - DenseNet121, a 2D-view encoder, a 3D decoder, and a 3D refiner}
\label{fig::fig1}
\end{figure}

3D coarse-to-fine transformer (3D-C2FT) consists of an image embedding module, a 2D-view encoder, a 3D decoder, and a 3D refiner, as illustrated in Fig. \ref{fig::fig1}. Specifically, the encoder accepts either single or multi-view images in the embedding form, which are managed by the image embedding module and then C2F patch attention is used to aggregate the different views of inputs by extracting feature representations for the decoder to reconstruct 3D volume. Finally, the refiner with the C2F cube attention mechanism is to rectify the defective reconstructed 3D volume. The 3D-C2FT network is explained in detail in the following subsections.

\subsection{Image Embedding Module}
\label{subsec::subsec31}
We leverage DenseNet121 \cite{Huang2017} as an image embedding module for the proposed 3D-C2FT. Given a set of $n$ view images of a 3D object $X = \{ X^{1}, X^{2}, \cdots, X^{n} \}$, each view image $X^{n}$ is fed into the image embedding module same as ViT \cite{Dosovitskiy2021} to obtain patch embeddings $P^{n} \in \mathbb{R}^{1 \times d}$, where $d$ is the embedding dimension. We follow the ViT setting where $d$ is fixed at 768.

\subsection{2D-view Encoder}
\label{subsec::subsec32}
The encoder of 3D-C2FT receives patch embeddings of view $n$, $P^{n}$ as well as its associated positional embeddings. These patches and positional embeddings are sent to a C2F patch attention block that consists of $I$-layer of MSA \cite{Vaswani2017} and multilayer perceptron (MLP), which is simply a fully connected feed forward network, as shown in Fig. \ref{fig::fig1}. The C2F patch attention block is repeated $J$ times. Thus, the output of the C2F patch attention block, $E^{n}_{I,j}$ is calculated as follows:
\begin{equation}
\label{eq::eq1}
    \hat{E}^{n}_{i,j} = \text{MSA} ( \text{Norm} ( P^{n}_{i-1,j}) ) + P^{n}_{i-1,j},
\end{equation}
\begin{equation}
\label{eq::eq2}
    E^{n}_{i,j} = \text{MLP} ( \text{Norm} ( \hat{E}^{n}_{i,j} ) ),
\end{equation}
where $\text{Norm}(\cdot)$ denotes layer normalization, $i$ is the layer index of $E^{n}_{i,j}$ where $i = 1, \cdots,I$ and $I$ is empirically set to 4 in this paper. $j$ is the block index of $E^{n}_{i,j}$, where $j = 1, \cdots, J$.

Each C2F patch attention block is responsible to extract the features i.e., $E^{n}_{I,j}$ in a coarse to fine-grained manner by shrinking the dimension of $E^{n}_{I,j}$ in a ratio of two. Thereafter, we fuse each $E^{n}_{I,j}$ from $J$ blocks by concatenation to produce a C2F patch embedding, $F^{n} \in \mathbb{R}^{1 \times [d+d/2+, \cdots,+d/2^{(J-1)} ]}$. Since $d$=768 and due to model size and computational complexity concerns, we set $J$=3 as shown in supplementary materials Section 6.1. The C2F patch embedding $F^{n}$ can be determined as follows:
\begin{equation}
\label{eq::eq3}
    F^{n} = \text{Norm} ( [ E^{n}_{I,1}, E^{n}_{I,2}, \cdots, E^{n}_{I,J} ] ),
\end{equation}
where $[\cdot]$ refers to the concatenation operator.

\subsection{3D Decoder}
\label{subsec::subsec33}
As shown in Fig. \ref{fig::fig1}, the 3D decoder learns and explores the global correlation between the multi-view images for reconstruction. The decoder only consists of a single 3D reconstruction block which is crucial for integrating and converting $F^{n}$ to a 3D volume $D$ with size $32 \times 32 \times 32$. The attention module of the 3D reconstruction block accepts both view interpretation matrix $H$ and $F^{n}$ to project different views of $F^{n}$ into the $D$. 
Specifically, given $F^{n}$ and $H \in \mathbb{R}^{g \times [d+d/2+, \cdots, +d/2^{(J-1)} ]}$, the decoder can easily aggregate the $n$-view of $F^{n}$ together and share across all potential inputs in the network. Here, $H$ is a randomly initialized learnable weight matrix that plays a role in interpreting and addressing $F^{n}$, and $g$ denotes the number of cube embeddings, with size $4 \times 4 \times 4$, that are required to assemble the $D$. In this paper, we set $g$=$8 \times 8 \times 8$ = 512. The 3D volume output of the decoder, $D$, is specified as:
\begin{equation}
\label{eq::eq4}
    D = \sigma \left ( \text{MLP} ( \text{MSA} (H, F^{n}) ) \right ) \in \mathbb{R}^{32 \times 32 \times 32},
\end{equation}
where $\sigma(\cdot)$ is a sigmoid activation function.
However, the reconstructed 3D volume, $D$ at this stage is defective (Section 4.3) due to simplicity of the decoder. Therefore, a refiner is needed for further rectification.  

\subsection{3D Refiner}
\label{subsec::subsec34}
We propose a C2F cube attention mechanism intending to correct and refine the $D$ surface for the refiner. The refiner is composed of $L$ C2F cube attention blocks where each block consists of $K$ layers of ViT-like attention module (Fig \ref{fig::fig1}). Here, we set $K$=6, which the experiments determine. In addition, $L$=2 is set in this work as the reconstructed volume resolution is a mere $32 \times 32 \times 32$. Therefore, a larger $L$ can be used for high-resolution volume.   

Specifically, $D$ from Eq. \ref{eq::eq4} is transformed to the C2F cube embeddings $\hat{R}_{k,l}$ by partitioning the $D$ to $\hat{R}_{k,1}$ with a size of $8 \times 8 \times 8$ in the first C2F cube attention block. For the second block, $\hat{R}_{k,2}$ size is further reduced by a factor of two; thus $4 \times 4 \times 4$. For each block, the cube attention block output can be computed from:

\begin{equation}
\label{eq::eq5}
    \hat{R}_{k,l} = \text{MSA} ( \text{Norm} ( \hat{R}_{k-1,l}) ) + \hat{R}_{k-1,l},
\end{equation}
\begin{equation}
\label{eq::eq6}
    R_{k,l} = \text{MLP} ( \text{Norm} ( \hat{R}_{k,l} ) ) + \hat{R}_{k,l}
\end{equation} 
where $k$ is the layer index of $R_{k,l}$ where $k = 1, 2, \cdots, K$ and $l$ is the block index of $R_{k,l}$ where $l = 1, 2, \cdots, L$.

By adopting the C2F notion, the proposed refiner leverages the C2F cube attention blocks of different scales of embeddings that can benefit the structure and parts correction of the 3D object. Furthermore, it enables the refiner to iteratively draw attention from coarse to temperate regions by rectifying the spatial surfaces gradually through multi-scale attention.

\subsection{Loss Function}
\label{subsec::subsec35}
For training, we adopt a combination of mean-square-error ($\mathcal{L}_{\text{MSE}}$) loss and 3D structural similarity ($\mathcal{L}_{\text{3D-SSIM}}$) loss as a loss function ($\mathcal{L}_{\text{total}}$) to capture better error information about the quality degradation of the 3D reconstruction. The motivation of adopting $\mathcal{L}_{\text{3D-SSIM}}$ loss \cite{Wang2004} is for 3D-C2FT to learn and produce visually pleasing 3D objects by quantifying volume structural differences between a reference and a reconstructed object. Furthermore, the $\mathcal{L}_{\text{MSE}}$ is used to evaluate the voxel loss that measures the L2 differences between the ground-truth (GT) and reconstructed objects. Thus, $\mathcal{L}_{\text{total}}$ is given as follows:
\begin{equation}
\label{eq::eq7}
    \mathcal{L}_{\text{total}}(Y, \hat{Y}) =  \mathcal{L}_{\text{MSE}}(Y, \hat{Y}) + \mathcal{L}_{\text{3D-SSIM}}(Y, \hat{Y}),
\end{equation}
\begin{equation}
\label{eq::eq8}
    \mathcal{L}_{\text{MSE}}(Y, \hat{Y}) = \frac{1}{M^3} \sum_{x=0}^{M} \sum_{y=0}^{M} \sum_{z=0}^{M}(Y_{x,y,z} - \hat{Y}_{x,y,z})^{2},
\end{equation}
\begin{equation}
\label{eq::eq9}
    \mathcal{L}_{\text{3D-SSIM}}(Y, \hat{Y}) = 1 - \frac{(2\mu_{Y}\mu_{\hat{Y}}+c_{1})/(2\sigma_{Y\hat{Y}}+c_{2})}{(\mu^{2}_{Y}+\mu^{2}_{\hat{Y}}+c_{1})(\sigma^{2}_{Y}+\sigma^{2}_{\hat{Y}}+c_{2})},
\end{equation}
where $Y$ is the GT of 3D volume and $\hat{Y}$ is reconstructed 3D volume. Note that, in Eq. \ref{eq::eq8}, $M$ refers to the dimension of 3D volume; $x$, $y$ and $z$ are defined as the indexes of voxel coordinates, respectively. In Eq. \ref{eq::eq9}, $\mu_{Y}$ and $\mu_{\hat{Y}}$ denote the mean of the voxels; $\sigma^{2}_{Y}$ and $\sigma^{2}_{\hat{Y}}$ are the variance of the voxels, respectively. In addition, $\sigma_{Y\hat{Y}}$ denotes the covariance between $Y$ and $\hat{Y}$. We set $c_{1}=0.01$ and $c_{2}=0.03$ to avoid instability when the mean and variances are close to zero.

\section{Experiments}
\label{sec::sec4}

\subsection{Evaluation Protocol and Implementation Details}
\label{subsec::subsec41}

\subsubsection{Dataset}
To have a fair comparison, we follow the protocols specified by \cite{Xie2020} and \cite{Yagubbayli2021} evaluate the proposed model. A subset of the ShapeNet dataset, which comprises 43,783 objects from 13 categories, is adopted. Each object is rendered from 24 different views. We resize the original images from $137 \times 137$ to $224 \times 224$, and the uniform background color is applied before passing them to the network. For each 3D object category, three subsets with a ratio of 70:10:20 for training, validation, and testing are partitioned randomly.

\subsubsection{Evaluation Metrics}
Intersection over Union (IoU) and F-score are adopted to evaluate the reconstruction performance of the proposed model. Provided predicted and GT 3D volume, the IoU score is defined as a ratio of voxels intersection of both volumes to their union \cite{Xie2020}; a higher IoU score implies a better reconstruction. F-score is used to quantify the object reconstruction quality. It measures the percentage of points from the object that are closer than a predetermined threshold \cite{Tatarchenko2019}. A higher F-score value suggests a better reconstruction. Therefore, the F-score@1\% is adopted in this paper.

\subsubsection{Implementation}
The proposed model is trained with the batch size 32 with view input images of size $224 \times 224$ and the dimension of 3D volume is set to $32 \times 32 \times 32$. For multi-view training, the number of input views is fixed to 8, which is the best obtained from the experiments. The full-fledged results are given in Section 4.3. 

During training, the views are randomly sampled out of 24 views at each iteration, and the network is trained by an SGD optimizer with an initial learning rate of 0.01. The learning rate decay is used, and it is subsequently reduced by $10^{-1}$ for every 500 epochs. The minimum learning rate is defined as $1.0 \times 10^{-4}$. Our network is implemented by using the PyTorch\footnote{PyTorch URL: https://pytorch.org/.} toolkit, and it is performed by an NVidia Tesla V100. The source code will be made publicly available.

\subsection{Result}
\label{subsec::subsec42}

\subsubsection{Multi-view 3D Reconstruction}
The performance comparisons of 3D-C2FT against several  benchmark models, namely 3D-R2N2 (RNN-based) \cite{Choy2016}, AttSets (CNN-based) \cite{Yang2020}, Pix2Vox++/F (CNN-based) \cite{Xie2020}, Pix2Vox++/A (CNN-based) \cite{Xie2020}, LegoFormer (transformer-based) \cite{Yagubbayli2021}, VoIT+ (transformer-based) \cite{Wang2021}, and EVoIT (transformer-based) \cite{Wang2021} are presented in this subsection. Table \ref{tab::tab1} shows reconstruction performance over a different number of views in terms of IoU and F-score on the ShapeNet dataset.

\newcolumntype{C}[1]{>{\centering\arraybackslash}p{#1}}
\begin{table}[!t]
\renewcommand{\arraystretch}{1.2}
\centering
\caption{Performance comparisons of single and multi-view 3D reconstruction on ShapeNet with IoU and F-score. The best score for each view is written in bold. * The `-' results of VoIT+ and EVoIT are not provided in \cite{Wang2021} and source code is not publicly available}
\label{tab::tab1}
\fontsize{7.5}{8}\selectfont
\begin{tabular}{cC{0.9cm}C{0.9cm}C{0.9cm}C{0.9cm}C{0.9cm}C{0.9cm}C{0.9cm}C{0.9cm}C{0.9cm}C{0.9cm}}
\hline
\multicolumn{1}{c|}{\multirow{2}{*}{\textbf{Model}}} & \multicolumn{9}{c}{\textbf{Number of views}} \\ \cline{2-10}
\multicolumn{1}{c|}{} & 1 & 2 & 3 & 4 & 5 & 8 & 12 & 18 & 20 \\ \hline
\multicolumn{1}{l|}{\textbf{\textit{Metric: IoU}}} & \multicolumn{9}{c}{} \\ \hline
\multicolumn{1}{l|}{\textbf{RNN \& CNN-based \:}} & \multicolumn{9}{c}{} \\
\multicolumn{1}{l|}{3D-R2N2 (2016)} & 0.560 & 0.603 & 0.617 & 0.625 & 0.634 & 0.635 & 0.636 & 0.636 & 0.636 \\
\multicolumn{1}{l|}{AttSets (2020) \:} & 0.642 & 0.662 & 0.670 & 0.675 & 0.677 & 0.685 & 0.688 & 0.692 & 0.693 \\
\multicolumn{1}{l|}{Pix2Vox++/F (2019)} & 0.645 & 0.669 & 0.678 & 0.682 & 0.685 & 0.690 & 0.692 & 0.693 & 0.694 \\
\multicolumn{1}{l|}{Pix2Vox++/A (2020) } & \textbf{0.670} & \textbf{0.695} & \textbf{0.704} & \textbf{0.708} & \textbf{0.711} & 0.715 & 0.717 & 0.718 & 0.719 \\ \hline
\multicolumn{1}{l|}{\textbf{Transformer-based}} & \multicolumn{9}{c}{} \\ 
\multicolumn{1}{l|}{LegoFormer (2021) \:} & 0.519 & 0.644 & 0.679 & 0.694 & 0.703 & 0.713 & 0.717 & 0.719 & 0.721 \\
\multicolumn{1}{l|}{VoIT+* (2021)} & - & - & - & 0.695 & - & 0.707 & 0.714 & - & 0.715 \\
\multicolumn{1}{l|}{EVoIT* (2021)} & - & - & - & 0.609 & - & 0.698 & \textbf{0.720} & - & \textbf{0.738} \\ \hline
\multicolumn{1}{l|}{3D-C2FT \:} & 0.629 & 0.678 & 0.695 & 0.702 & 0.708 & \textbf{0.716} & \textbf{0.720} & \textbf{0.723} & 0.724 \\
\hline
\multicolumn{1}{l|}{\textbf{\textit{Metric: F-score \:}}} & \multicolumn{9}{c}{} \\ \hline
\multicolumn{1}{l|}{\textbf{RNN \& CNN-based}} & \multicolumn{9}{c}{} \\
\multicolumn{1}{l|}{3D-R2N2 (2016)} & 0.351 & 0.368 & 0.372 & 0.378 & 0.382 & 0.383 & 0.382 & 0.382 & 0.383 \\
\multicolumn{1}{l|}{AttSets (2020) } & 0.395 & 0.418 & 0.426 & 0.430 & 0.432 & 0.444 & 0.445 & 0.447 & 0.448 \\
\multicolumn{1}{l|}{Pix2Vox++/F (2019)} & 0.394 & 0.422 & 0.432 & 0.437 & 0.440 & 0.446 & 0.449 & 0.450 & 0.451 \\
\multicolumn{1}{l|}{Pix2Vox++/A (2020)} & \textbf{0.436} & \textbf{0.452} & \textbf{0.455} & \textbf{0.457} & \textbf{0.458} & 0.459 & 0.460 & 0.461 & 0.462 \\ \hline
\multicolumn{1}{l|}{\textbf{Transformer-based}} & \multicolumn{9}{c}{} \\
\multicolumn{1}{l|}{LegoFormer (2021) \:} & 0.282 & 0.392 & 0.428 & 0.444 & 0.453 & 0.464 & 0.470 & 0.472 & 0.473 \\
\multicolumn{1}{l|}{VoIT+* (2021)} & - & - & - & 0.451 & - & 0.464 & 0.469 & - & 0.474 \\
\multicolumn{1}{l|}{EVoIT* (2021)} & - & - & - & 0.358 & - & 0.448 & \textbf{0.475} & - & \textbf{0.497} \\ \hline
\multicolumn{1}{l|}{3D-C2FT \:} & 0.371 & 0.424 & 0.443 & 0.452 & \textbf{0.458} & \textbf{0.468} & \textbf{0.475} & \textbf{0.477} & 0.479 \\
\hline
\end{tabular}
\end{table}

Compared with RNN-based, and CNN-based models, it can be observed that 3D-C2FT achieves the highest IoU and F-score with more than 8 views. As an illustration, we demonstrate several reconstruction instances from the ShapeNet dataset as shown in supplementary materials Section 6.2. With more than 8 views input, the objects reconstructed by the 3D-C2FT have complete and smoother surfaces, even compared with the LegoFormer. These results suggest that C2F patch and cube attention mechanisms play a crucial role in multi-view 3D object reconstruction. On the other hand, Pix2Vox++/A is the best performing model with less than five views compared to transformer-based models, although the difference with 3D-C2FT in terms of IoU and F-score is not significant. However, when given large number of views, it fails to exploit the relations of multi-view features.

Among transformer-based models, 3D-C2FT outperforms LegoFormer, VoIT+ and EVoIT significantly almost all views in terms of IoU score and F-score, even under extreme cases where one view image is used as input. Note VoIT+ and EVoIT are published without code availability; thus, we only report the results for 4, 8, 12, and 20 views inputs \cite{Wang2021}. Nevertheless, the results reveal the distinctive advantages of the C2F patch and cube attention over competing transformer models that utilize a plain single-scaled MSA mechanism in both encoder and decoder. However, 3D-C2FT underperforms EVoIT for 20 views. We speculate that the issue is associated with the attention layer in the decoder being less effective for a large number of input views.

\begin{figure}[!t]
\centering
\includegraphics[width=0.80\textwidth]{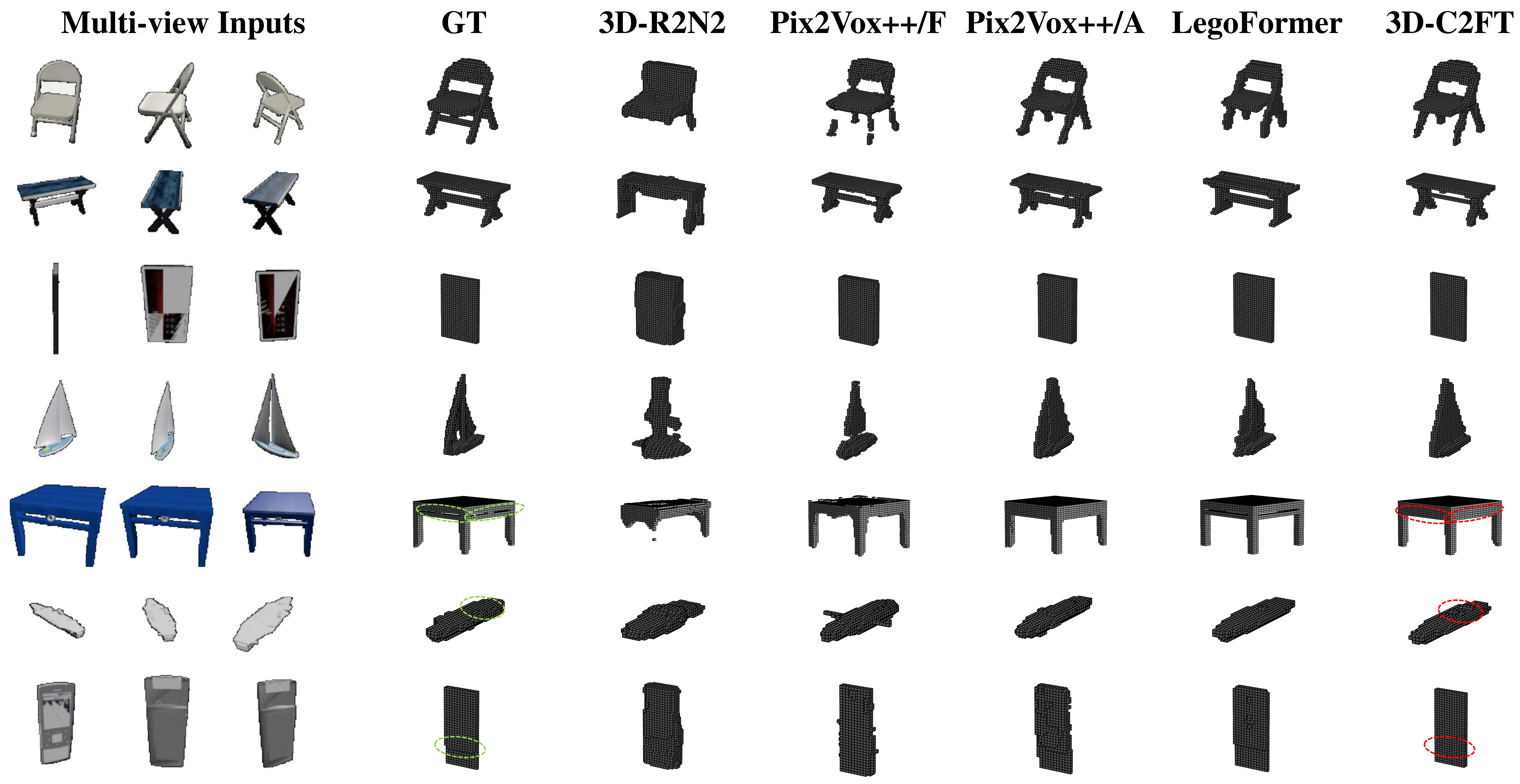}
\caption{3D object reconstruction using 8 views (only 3 are shown) for specific categories: \textit{chair}, \textit{table}, \textit{telephone} and \textit{table}}
\label{fig::fig2}
\end{figure}

Next, we analyze a few poor reconstructed categories by the 3D-C2FT as shown in Fig. \ref{fig::fig2} and supplementary materials Section 6.3. As shown in Fig. \ref{fig::fig2}, row \#5--\#7, 3D-C2FT fails to reconstruct a few specific regions marked with red dotted circles. This suggests that the C2F patch attention mechanism could be affected by the presence of view images with similar view orientations.

\subsubsection{Performance Comparisons with Occlusion Images} 
In this subsection, the reconstruction performance of 3D-C2FT and the competing models is evaluated with occluded images. Here, we use the testing set from the subset of the ShapeNet dataset by following the same protocols in the previous section. We add occlusion boxes with different sizes to the odd number ordered lists of 2D view images: \textit{10}$\times$\textit{10}, \textit{15}$\times$\textit{15}, \textit{20}$\times$\textit{20}, \textit{25}$\times$\textit{25}, \textit{30}$\times$\textit{30}, \textit{35}$\times$\textit{35} and \textit{40}$\times$\textit{40}, which impeded the essential parts of the images intentionally. Readers are referred to the supplementary materials Section 6.4 for details.

As shown in Table \ref{tab::tab2}, the proposed 3D-C2FT achieves the best IoU scores for all sizes of occlusion boxes. Among the benchmark models, Pix2Vox++/A performs second-best over various occlusion boxes. Surprisingly, LegoFormer unperformed with IoU scores between 0.678 and 0.698 only. Unlike the LegoFormer that draws attention to the fine-grained features only to explore the correlation between multi-view images, 3D-C2FT focuses on both coarse and fine-grained features so that they can assist the model in exploiting global and local interactions within the occluded regions, which make the model more robust against predicting unknown elements in the multi-view images.

\begin{table}[!t]
\renewcommand{\arraystretch}{1.2}
\centering
\caption{Performance comparisons of 12-view 3D reconstruction on ShapeNet using different sizes of occlusion box. The best score for each size is highlighted in bold}
\label{tab::tab2}
\fontsize{7.5}{8}\selectfont
\begin{tabular}{c|C{1.2cm}C{1.2cm}C{1.2cm}C{1.2cm}C{1.2cm}C{1.2cm}C{1.2cm}}
\hline
\multicolumn{1}{c|}{\multirow{2}{*}{\textbf{Model}}} & \multicolumn{7}{c}{\textbf{Occlusion Box}} \\ \cline{2-8}
\multicolumn{1}{c|}{} & \textit{10}$\times$\textit{10} & \textit{15}$\times$\textit{15} & \textit{20}$\times$\textit{20} & \textit{25}$\times$\textit{25} & \textit{30}$\times$\textit{30} & \textit{35}$\times$\textit{35} & \textit{40}$\times$\textit{40} \\ \hline
\multicolumn{1}{l|}{\textbf{\textit{Metric: IoU}}} & \multicolumn{7}{c}{} \\ \hline
\multicolumn{1}{l|}{Pix2Vox++/F \:} & 0.665 & 0.663 & 0.663 & 0.662 & 0.661 & 0.660 & 0.660 \\
\multicolumn{1}{l|}{Pix2Vox++/A \:} & 0.709 & 0.706 & 0.704 & 0.700 & 0.697 & 0.692 & 0.688 \\
\multicolumn{1}{l|}{LegoFormer \:} & 0.698 & 0.696 & 0.694 & 0.689 & 0.686 & 0.683 & 0.678 \\
\multicolumn{1}{l|}{3D-C2FT \:} & \textbf{0.712} & \textbf{0.710} & \textbf{0.708} & \textbf{0.703}& \textbf{0.700} & \textbf{0.694} & \textbf{0.692} \\
\hline
\multicolumn{1}{l|}{\textbf{\textit{Metric: F-score \:\:}}} & \multicolumn{7}{c}{} \\ 
\hline
\multicolumn{1}{l|}{Pix2Vox++/F \:} & 0.386 & 0.384 & 0.383 & 0.382 & 0.382 & 0.381 & 0.381 \\
\multicolumn{1}{l|}{Pix2Vox++/A \:} & 0.462 & 0.460 & 0.458 & 0.454 & \textbf{0.451} & 0.442 & 0.439 \\
\multicolumn{1}{l|}{LegoFormer \:} & 0.448 & 0.446 & 0.445 & 0.441 & 0.438 & 0.436 & 0.431 \\
\multicolumn{1}{l|}{3D-C2FT \:} & \textbf{0.464} & \textbf{0.462} & \textbf{0.459} & \textbf{0.455} & \textbf{0.451} & \textbf{0.446} & \textbf{0.443} \\
\hline
\end{tabular}
\end{table}

\subsubsection{Experiments on Multi-view Real-life Dataset}
This subsection presents the performance comparisons with Multi-view Real-life Dataset. The dataset contains single or multi-view images that follow the object categories of the ShapeNet dataset. To create our dataset, we first select the category name and search with Google image search engine. Then, we verify the search result to ensure that each image is correctly labeled with at least single or multi-view images depending on the available combinations. The dataset is designed as a testing set that only contains 60 samples across 13 categories with three cases:
\begin{itemize}
\item \textit{Case I}: samples with at least 1 to 5 views;
\item \textit{Case II}: samples with at least 6 to 11 views;
\item \textit{Case III}: samples with at least 12 or above views;
\end{itemize}

\begin{figure}[!t]
\centering
\includegraphics[width=0.70\textwidth]{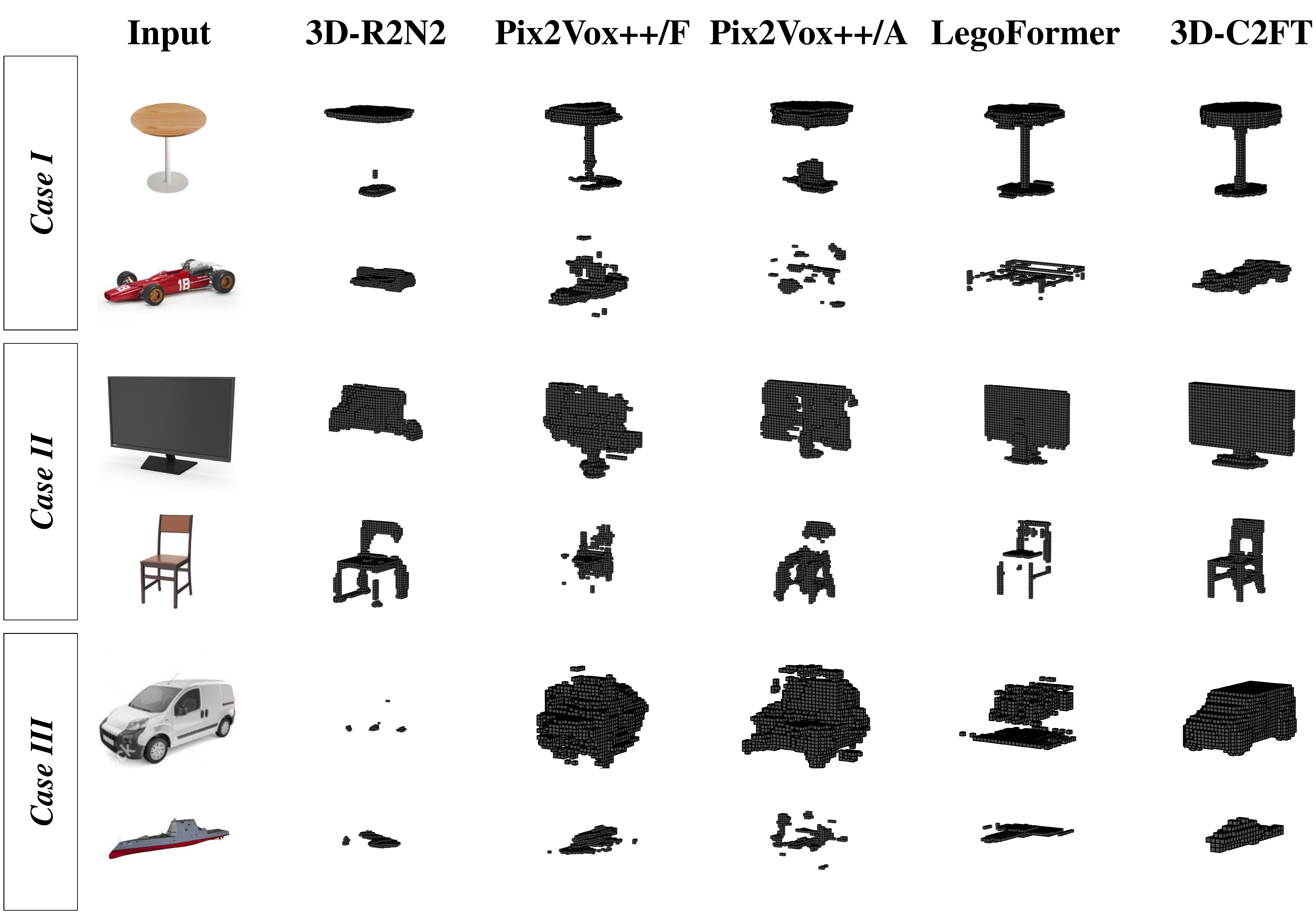}
\caption{3D object reconstruction for \textit{Case I}, \textit{Case II} and \textit{Case III} with Multi-view Real-life dataset}
\label{fig::fig3}
\end{figure}

The experiment is designed to verify the 3D objects based on human perception without the GT of 3D volume. In Fig. \ref{fig::fig3} and supplementary materials Section 6.5, we show the qualitative results for single and multi-view 3D reconstruction. It is observed that 3D-C2FT performs substantially better than the benchmark models in terms of the refined 3D volume and surface quality even in \textit{Case I}. Interestingly, most benchmark models fail to reconstruct the 3D objects, or the surface is not accurately generated. This is likely because the benchmark models lack in utilizing multi-scale feature extraction mechanisms, which causes these models to fail to capture relevant components more richly. 

Note that  \cite{Yagubbayli2021} advocates a part-by-part composition notion by means of fine-to-fine attention, which is opposed to our global-to-local attention notion. As depicted in Fig. \ref{fig::fig3}, LegoFormer underperforms the proposed model. This could be due to LegoFormer does not utilize 3D structure information essential in interpreting global interaction between semantic components. As a result, the model ignores the importance of the structure and orientation information of the 3D objects in guiding the model to perform reconstruction in challenging environments.

The advantage of the C2F attention mechanism is that it fully utilizes the coarse and fine-grained features to be paired and represents the specific information flow that benefits the 3D objects reconstruction task.

\subsubsection{Model Complexity} Table \ref{tab::tab3} tabulates the number of parameters in various benchmark models. Note that the CNN backbone refers to the pre-trained image embedding module. 3D-C2FT has a smaller number of parameters than most of the competing models. Although both 3D-R2N2 and AttSets have significantly smaller parameter sizes, their reconstruction performances are below par, as revealed in Table \ref{tab::tab1}.

\begin{table}[!t]
\renewcommand{\arraystretch}{1.2}
\centering
\caption{Comparisons on parameter sizes of competing models}
\label{tab::tab3}
\fontsize{7.5}{8}\selectfont
\begin{tabular}{ccc}
\hline
\textbf{Model} & \: \textbf{Param. (M)} \: & \: \textbf{CNN Backbone} \: \\
\hline
\multicolumn{1}{l}{3D-R2N2} & 36.0 & - \\
\multicolumn{1}{l}{AttSets} & 53.1 & - \\
\multicolumn{1}{l}{Pix2Vox++/F}\: & 114.2 & ResNet50 \\
\multicolumn{1}{l}{Pix2Vox++/A}\: & 96.3 & ResNet50 \\
\multicolumn{1}{l}{LegoFormer} & 168.4 & VGG16 \\
\multicolumn{1}{l}{3D-C2FT} & 90.1 & DenseNet121 \\
\hline
\end{tabular}
\end{table}

\subsection{Ablation Study}
\label{subsec::subsec43}
In this Section, all the experiments are conducted with the ShapeNet dataset described in Section \ref{subsec::subsec41}.

\subsubsection{Visualization on C2F Attention}
\begin{figure}
\centering
\includegraphics[width=0.55\textwidth]{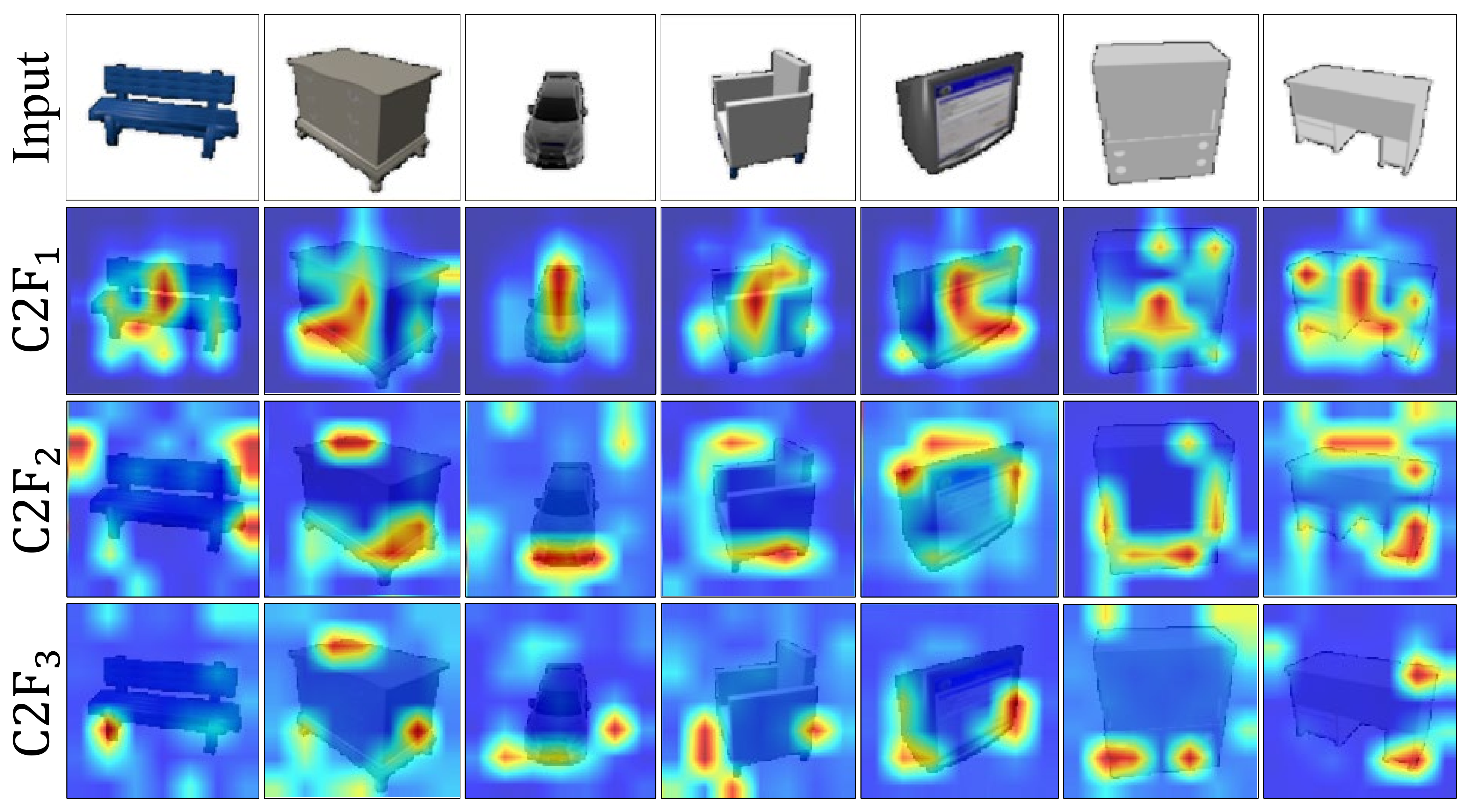}
\caption{Visualization of attention maps for each C2F patch attention block. This figure is best viewed on a screen}
\label{fig::fig4}
\end{figure}

To better understand the benefit of C2F attention, we present visualization results of three C2F patch attention blocks separately, labeled as $\text{C2F}_{1}$, $\text{C2F}_{2}$ and $\text{C2F}_{3}$ with attention rollout \cite{Abnar2020}. Fig. \ref{fig::fig4} depicts the attention maps of each stage of the C2F patch attention block, which interpret and focus semantically relevant regions that result in a coarse to fine-grained manner. 

As shown in Fig. \ref{fig::fig4}, the $\text{C2F}_{1}$ attention is drawn to the central parts of the given images, which represent coarse or global features of the 3D objects. Similarly, the $\text{C2F}_{2}$ shows attention is drawn toward the edges of the objects, which can assist the model to distinguish the view orientations for reconstruction. The final block ($\text{C2F}_{3}$) suggests that the attention is drawn to the specific parts of the objects, which can be considered as fine-grained features. This facilitates the model to capture the 3D surface details. The results indicate that the proposed coarse-to-fine mechanism pays specific attention to the relevant components in the 3D objects, which is essential for the decoder to predict and reconstruct accurate 3D volume.

\subsubsection{Reconstruction with and without Refiner}
\begin{figure}[!t]
\centering
\includegraphics[width=0.70\textwidth]{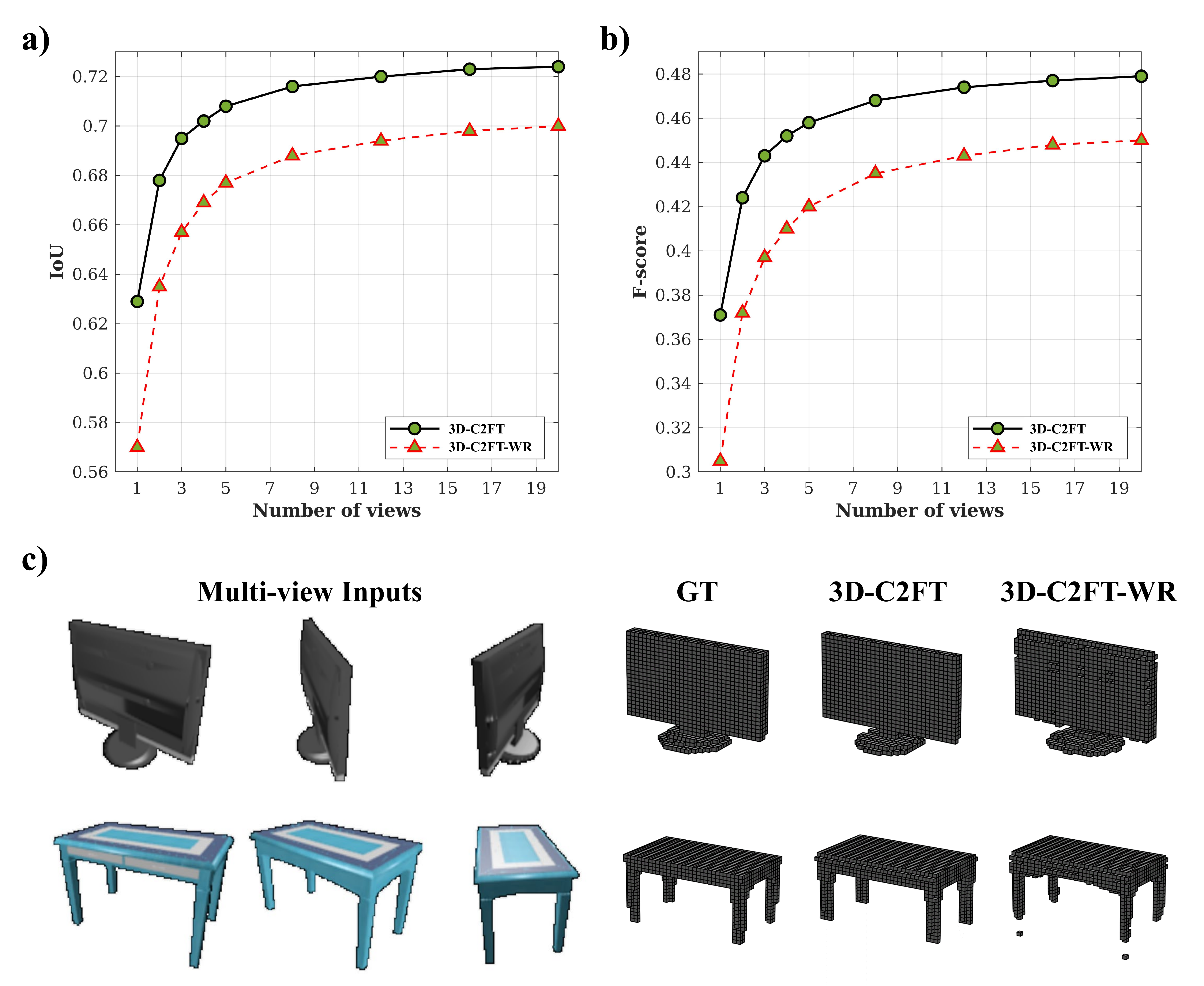}
\caption{Performance comparisons of multi-view 3D reconstruction results on 3D-C2FT and 3D-C2FT-WR. a) IoU score; b) F-score; c) 3D reconstruction results using 8 views (only 3 are shown) between 3D-C2FT and 3D-C2FT-WR. This figure is best viewed on a screen}
\label{fig::fig5}
\end{figure}

In Fig. \ref{fig::fig5}a and \ref{fig::fig5}b, we evaluate the influence of the refiner on 3D reconstruction results using 3D-C2FT and 3D-C2FT without refiner (3D-C2FT-WR). We observe that 3D-C2FT can significantly achieve the best IoU score and F-score in all views. We also show several qualitative 3D reconstruction results in Fig. \ref{fig::fig5}c. Without refiner, the decoder generates defective 3D surfaces due to a lack of drawing attention to rectifying wrongly reconstructed parts within the 3D objects. Therefore, our analysis indicates that the refiner plays a crucial role in improving the 3D surface quality and can remove the noise of the reconstructed 3D volume from the decoder.

\subsubsection{Loss Functions} 
We also show the performance of 3D-C2FT trained with different loss functions, namely $\mathcal{L}_{\text{total}}$, $\mathcal{L}_{\text{MSE}}$ and $\mathcal{L}_{\text{3D-SSIM}}$, as shown in Table \ref{tab::tab4}. It is noticed that 3D-C2FT trained with $\mathcal{L}_{\text{total}}$ achieves the best IoU and F-score compared to either $\mathcal{L}_{\text{MSE}}$ or $\mathcal{L}_{\text{3D-SSIM}}$ alone. $\mathcal{L}_{\text{MSE}}$ minimizes the voxel grid error between the predicted and GT of 3D objects. In contrast, $\mathcal{L}_{\text{3D-SSIM}}$ reduces the structural difference, which benefits in improving the surface of predicted 3D objects. As a result, both loss functions drive the model training towards more reliable and better surface quality reconstruction.

\begin{table}[!t]
\renewcommand{\arraystretch}{1.2}
\centering
\caption{Performance comparisons of multi-view 3D reconstruction results on 3D-C2FT using different loss functions. The best score for each view is written in bold}
\label{tab::tab4}
\fontsize{7.5}{8}\selectfont
\begin{tabular}{cC{0.9cm}C{0.9cm}C{0.9cm}C{0.9cm}C{0.9cm}C{0.9cm}C{0.9cm}C{0.9cm}C{0.9cm}C{0.9cm}}
\hline
\multicolumn{1}{c|}{\multirow{2}{*}{\textbf{Loss function}}} & \multicolumn{9}{c}{\textbf{Number of views}} \\ \cline{2-10}
\multicolumn{1}{c|}{} & 1 & 2 & 3 & 4 & 5 & 8 & 12 & 18 & 20 \\ \hline
\multicolumn{1}{l|}{\textbf{\textit{Metric: IoU}}} & \multicolumn{9}{c}{} \\ \hline
\multicolumn{1}{l|}{$\mathcal{L}_{\text{MSE}}$ \:} & 0.623 & 0.673 & 0.683 & 0.696 & 0.701 & 0.708 & 0.713 & 0.716 & 0.716 \\
\multicolumn{1}{l|}{$\mathcal{L}_{\text{3D-SSIM}}$ \:} & 0.617 & 0.668 & 0.683 & 0.691 & 0.696 & 0.705 & 0.709 & 0.712 & 0.712 \\
\multicolumn{1}{l|}{$\mathcal{L}_{\text{total}}$ \:} & \textbf{0.629} & \textbf{0.678} & \textbf{0.695} & \textbf{0.702} & \textbf{0.708} & \textbf{0.716} & \textbf{0.720} & \textbf{0.723} & \textbf{0.724} \\
\hline
\multicolumn{1}{l|}{\textbf{\textit{Metric: F-score \:}}} & \multicolumn{9}{c}{} \\ \hline
\multicolumn{1}{l|}{$\mathcal{L}_{\text{MSE}}$ \:} & 0.369 & 0.421 & 0.437 & 0.451 & 0.457 & 0.466 & 0.471 & 0.474 & 0.476 \\
\multicolumn{1}{l|}{$\mathcal{L}_{\text{3D-SSIM}}$ \:} & 0.365 & 0.418 & 0.437 & 0.446 & 0.452 & 0.462 & 0.467 & 0.470 & 0.472 \\
\multicolumn{1}{l|}{$\mathcal{L}_{\text{total}}$ \:} & \textbf{0.371} & \textbf{0.424} & \textbf{0.443} & \textbf{0.452} & \textbf{0.458} & \textbf{0.468} & \textbf{0.475} & \textbf{0.477} & \textbf{0.479} \\
\hline
\end{tabular}
\end{table}

\subsubsection{Training with Different View Counts}
This ablation investigates the object reconstruction performance with respect to the number of input view images used in 3D-C2FT training. Here, we fix the view input numbers at 4, 8, and 12. In addition, the input views are randomly sampled at every training iteration. From Table \ref{tab::tab5}, it is interesting to see that the best number of input views for training is 8, but not the larger view count such as 12 by intuition. This phenomenon could be associated with the limitation of the decoder to aggregate coarse to fine-grained features with similar orientation views.

\begin{table}
\renewcommand{\arraystretch}{1.2}
\centering
\caption{Performance comparisons of single and multi-view 3D reconstruction on 3D-C2FT using specific view counts for training. The best score for each view is written in bold}
\label{tab::tab5}
\fontsize{7.5}{8}\selectfont
\begin{tabular}{cC{0.9cm}C{0.9cm}C{0.9cm}C{0.9cm}C{0.9cm}C{0.9cm}C{0.9cm}C{0.9cm}C{0.9cm}C{0.9cm}}
\hline
\multicolumn{1}{c|}{\multirow{2}{*}{\begin{tabular}[c]{@{}c@{}}\textbf{Training view}\\ \textbf{counts}\end{tabular}}} & \multicolumn{9}{c}{\textbf{Testing view counts}} \\ \cline{2-10}
\multicolumn{1}{c|}{} & 1 & 2 & 3 & 4 & 5 & 8 & 12 & 18 & 20 \\ 
\hline
\multicolumn{1}{l|}{\textbf{\textit{Metric: IoU}}} & \multicolumn{9}{c}{} \\ 
\hline
\multicolumn{1}{l|}{4 views \:} & \textbf{0.629} & 0.672 & 0.688 & 0.694 & 0.699 & 0.707 & 0.711 & 0.713 & 0.714 \\
\multicolumn{1}{l|}{8 views \:} & \textbf{0.629} & \textbf{0.678} & \textbf{0.695} & \textbf{0.702} & \textbf{0.708} & \textbf{0.716} & \textbf{0.720} & \textbf{0.723} & \textbf{0.724} \\
\multicolumn{1}{l|}{12 views \:} & 0.628 & \textbf{0.678} & \textbf{0.695} & \textbf{0.702} & \textbf{0.708} & \textbf{0.716} & \textbf{0.720} & \textbf{0.723} & \textbf{0.724} \\
\hline
\multicolumn{1}{l|}{\textbf{\textit{Metric: F-score \:}}} & \multicolumn{9}{c}{} \\ 
\hline
\multicolumn{1}{l|}{4 views \:} & \textbf{0.374} & 0.421 & 0.438 & 0.446 & 0.451 & 0.460 & 0.466 & 0.469 & 0.470 \\
\multicolumn{1}{l|}{8 views \:} & 0.371 & \textbf{0.424} & \textbf{0.443} & \textbf{0.452} & \textbf{0.458} & \textbf{0.468} & \textbf{0.475} & \textbf{0.477} & \textbf{0.479} \\
\multicolumn{1}{l|}{12 views \:} & 0.370 & 0.423 & 0.442 & \textbf{0.452} & \textbf{0.458} & \textbf{0.468} & \textbf{0.475} & \textbf{0.477} & \textbf{0.479} \\
\hline
\end{tabular}
\end{table}

\section{Conclusion}
\label{sec::sec5}
This paper proposed a multi-view 3D reconstruction model that employs a C2F patch attention mechanism in the 2D encoder, a 3D decoder, and a C2F cube attention mechanism in the refiner. Our experiments showed that 3D-C2FT could achieve significant results compared to the several competing models on the ShapeNet dataset. Further study with the Multi-view Real-life dataset showed that 3D-C2FT is far more robust than other models. For future works, we will consider improving the attention mechanism and the scaling of 3D-C2FT. Finally, we will explore other mechanisms to enhance the 3D reconstruction task, which we hope will facilitate the practical usage of robotics, historical artifacts, and other related domains.

%
%
\bibliographystyle{splncs04}
\bibliography{myref}

\begin{thebibliography}{10}
\providecommand{\url}[1]{\texttt{#1}}
\providecommand{\urlprefix}{URL }
\providecommand{\doi}[1]{https://doi.org/#1}

\bibitem{Abnar2020}
Abnar, S., Zuidema, W.: Quantifying attention flow in transformers. arXiv
  e-prints  (2020), \url{https://arxiv.org/abs/2005.00928}

\bibitem{Burchfiel2017}
Burchfiel, B., Konidaris, G.: {Bayesian eigenobjects: A unified framework for
  3D robot perception}. In: Robotics: Science and Systems. vol.~13 (2017)

\bibitem{Choy2016}
Choy, C.B., Xu, D., Gwak, J., Chen, K., Savarese, S.: {3D-R2N2: A unified
  approach for single and multi-view 3D object reconstruction}. In: European
  Conference on Computer Vision (ECCV). pp. 628--644 (2016)

\bibitem{Dosovitskiy2021}
Dosovitskiy, A., Beyer, L., Kolesnikov, A., Weissenborn, D., Zhai, X.,
  Unterthiner, T., Dehghani, M., Minderer, M., Heigold, G., Gelly, S.,
  Uszkoreit, J., Houlsby, N.: {An image is worth 16x16 words: Transformers for
  image recognition at scale}. In: International Conference on Learning
  Representations (ICLR) (2021)

\bibitem{Gao2014}
Gao, Y., Luo, J., Qiu, H., Wu, B.: {Survey of structure from motion}. In:
  Proceedings of 2014 International Conference on Cloud Computing and Internet
  of Things. pp. 72--76 (2014)

\bibitem{Groen2019}
Groen, I.I.A., Baker, C.I.: {Previews scenes in the human brain: Comparing 2D
  versus 3D representations}. Neuron  \textbf{101}(1),  8--10 (2019)

\bibitem{Han2021}
Han, X.F., Laga, H., Bennamoun, M.: {Image-based 3D object reconstruction:
  State-of-the-art and trends in the deep learning era}. IEEE Transactions on
  Pattern Analysis and Machine Intelligence  \textbf{43}(5),  1578--1604 (2021)

\bibitem{Huang2017}
Huang, G., Liu, Z., van~der Maaten, L., Weinberger, K.Q.: {Densely connected
  convolutional networks}. In: Proceedings of the IEEE Conference on Computer
  Vision and Pattern Recognition (CVPR). pp. 4700--4708 (2017)

\bibitem{Kar2017}
Kar, A., H\"ane, C., Malik, J.: {Learning a multi-view stereo machine}. In:
  Proceedings of the 31st International Conference on Neural Information
  Processing Systems (NIPS). p. 364–375. Curran Associates, Inc. (2017)

\bibitem{Kargas2019}
Kargas, A., Loumos, G., Varoutas, D.: {Using different ways of 3{D}
  reconstruction of historical cities for gaming purposes: The case study of
  Nafplio}. Heritage  \textbf{2}(3),  1799--1811 (2019)

\bibitem{Nabil2014}
Nabil, M., Saleh, F.: {3D reconstruction from images for museum artefacts: A
  comparative study}. In: International Conference on Virtual Systems and
  Multimedia (VSMM). pp. 257--260. IEEE (2014)

\bibitem{Nguyen2019}
Nguyen, T.Q., Salazar, J.: Transformers without tears: Improving the
  normalization of self-attention. In: Proceedings of the 16th International
  Conference on Spoken Language Translation. Hong Kong (2019)

\bibitem{Park2022}
Park, N., Kim, S.: How do vision transformers work? In: International
  Conference on Learning Representations (2022)

\bibitem{Pavaloiu2014}
Păvăloiu, I.B., Vasilăţeanu, A., Goga, N., Marin, I., Ilie, C., Ungar, A.,
  Pătraşcu, I.: {3D dental reconstruction from CBCT data}. In: International
  Symposium on Fundamentals of Electrical Engineering (ISFEE). pp.~4--9 (2014)

\bibitem{Roointan2019}
Roointan, S., Tavakolian, P., Sivagurunathan, K.S., Floryan, M., Mandelis, A.,
  Abrams, S.H.: {3D dental subsurface imaging using enhanced truncated
  correlation-photothermal coherence tomography}. Scientific Reports
  \textbf{9}(1),  1--12 (2019)

\bibitem{Shi2017}
Shi, Q., Li, C., Wang, C., Luo, H., Huang, Q., Fukuda, T.: {Design and
  implementation of an omnidirectional vision system for robot perception}.
  Mechatronics  \textbf{41},  58--66 (2017)

\bibitem{Silveira2008}
Silveira, G., Malis, E., Rives, P.: {An efficient direct approach to visual
  SLAM}. IEEE Transactions on Robotics  \textbf{24}(5),  969--979 (2008)

\bibitem{Tatarchenko2017}
Tatarchenko, M., Dosovitskiy, A., Brox, T.: Octree generating networks:
  Efficient convolutional architectures for high-resolution 3{D} outputs. In:
  IEEE International Conference on Computer Vision (ICCV). pp. 2088--2096
  (2017)

\bibitem{Tatarchenko2019}
Tatarchenko*, M., Richter*, S.R., Ranftl, R., Li, Z., Koltun, V., Brox, T.:
  What do single-view 3{D} reconstruction networks learn? In: Proceedings of
  the IEEE Conference on Computer Vision and Pattern Recognition (CVPR). pp.
  3405--3414 (2019)

\bibitem{Tron2014}
Tron, R., Vidal, R.: {Distributed 3-D localization of camera sensor networks
  from 2-D image Measurements}. IEEE Transactions on Automatic Control
  \textbf{59}(12),  3325--3340 (2014)

\bibitem{Vaswani2017}
Vaswani, A., Shazeer, N., Parmar, N., Uszkoreit, J., Jones, L., Gomez, A.N.,
  Kaiser, L.u., Polosukhin, I.: {Attention is all you need}. In: Proceedings of
  the 31st International Conference on Neural Information Processing Systems
  (NIPS). vol.~30, p. 6000–6010 (2017)

\bibitem{Wang2021}
Wang, D., Cui, X., Chen, X., Zou, Z., Shi, T., Salcudean, S., Wang, Z.J., Ward,
  R.: {Multi-view 3D reconstruction with transformer}. In: International
  Conference on Computer Vision (ICCV). pp. 5722--5731 (2021)

\bibitem{Wang2004}
Wang, Z., Bovik, A., Sheikh, H., Simoncelli, E.: Image quality assessment: from
  error visibility to structural similarity. IEEE Transactions on Image
  Processing  \textbf{13}(4),  600--612 (2004)

\bibitem{Wilson2014}
Wilson, K., Snavely, N.: {Robust global translations with 1DSfM}. In: European
  Conference on Computer Vision (ECCV). pp. 61--75 (2014)

\bibitem{Wu2015}
Wu, Z., Song, S., Khosla, A., Yu, F., Zhang, L., Tang, X., Xiao, J.: {3D
  ShapeNets: A deep representation for volumetric shapes}. In: Proceedings of
  the IEEE Conference on Computer Vision and Pattern Recognition (CVPR). pp.
  1912--1920 (2015)

\bibitem{Xie2019}
Xie, H., Yao, H., Sun, X., Zhou, S., Zhang, S.: {Pix2Vox: Context-aware 3D
  reconstruction from single and multi-view images}. In: IEEE International
  Conference on Computer Vision (ICCV). pp. 2690--2698 (2019)

\bibitem{Xie2020}
Xie, H., Yao, H., Zhang, S., Zhou, S., Sun, W.: {Pix2Vox++: Multi-scale
  context-aware 3D object reconstruction from single and multiple images}.
  International Journal of Computer Vision  \textbf{128}(12),  2919--2935
  (2020)

\bibitem{Yagubbayli2021}
Yagubbayli, F., Tonioni, A., Tombari, F.: {LegoFormer: Transformers for
  block-by-block multi-view 3D reconstruction}. arXiv e-prints  (2021),
  \url{http://arxiv.org/abs/2106.12102}

\bibitem{Yang2020}
Yang, B., Wang, S., Markham, A., Trigoni, N.: {Robust attentional aggregation
  of deep feature sets for multi-view 3D reconstruction}. International Journal
  of Computer Vision  \textbf{128}(1),  53--73 (2020)

\end{thebibliography}

\title{\text{Supplementary Materials of}\\[0.7cm] 3D-C2FT: Coarse-to-fine Transformer for Multi-view 3D Reconstruction} 

\titlerunning{3D-C2FT: Coarse-to-fine Transformer for Multi-view 3D Reconstruction}
%
\author{}
\authorrunning{L. Tiong et al.}
%
\institute{}
\maketitle
\section*{6 Appendix}
\label{sec::secAppendix}

\subsection*{6.1 Additional Ablation Study}

\subsubsection{Training with Different Number of C2F Patch Attention Block}
In Supplementary Fig.\ref{sfig::sfig1}, we visualize the performances on the ShapeNet validation dataset with respect to the number of C2F patch attention blocks, $J= {3, 4, 5}$ used in the proposed network. We observe that 3D-C2FT with $J$=3 achieves the smallest loss. In the main text, this justifies $J$ = 3 is set for 3D-C2FT.

\begin{figure}
\centering
\includegraphics[width=0.80\textwidth]{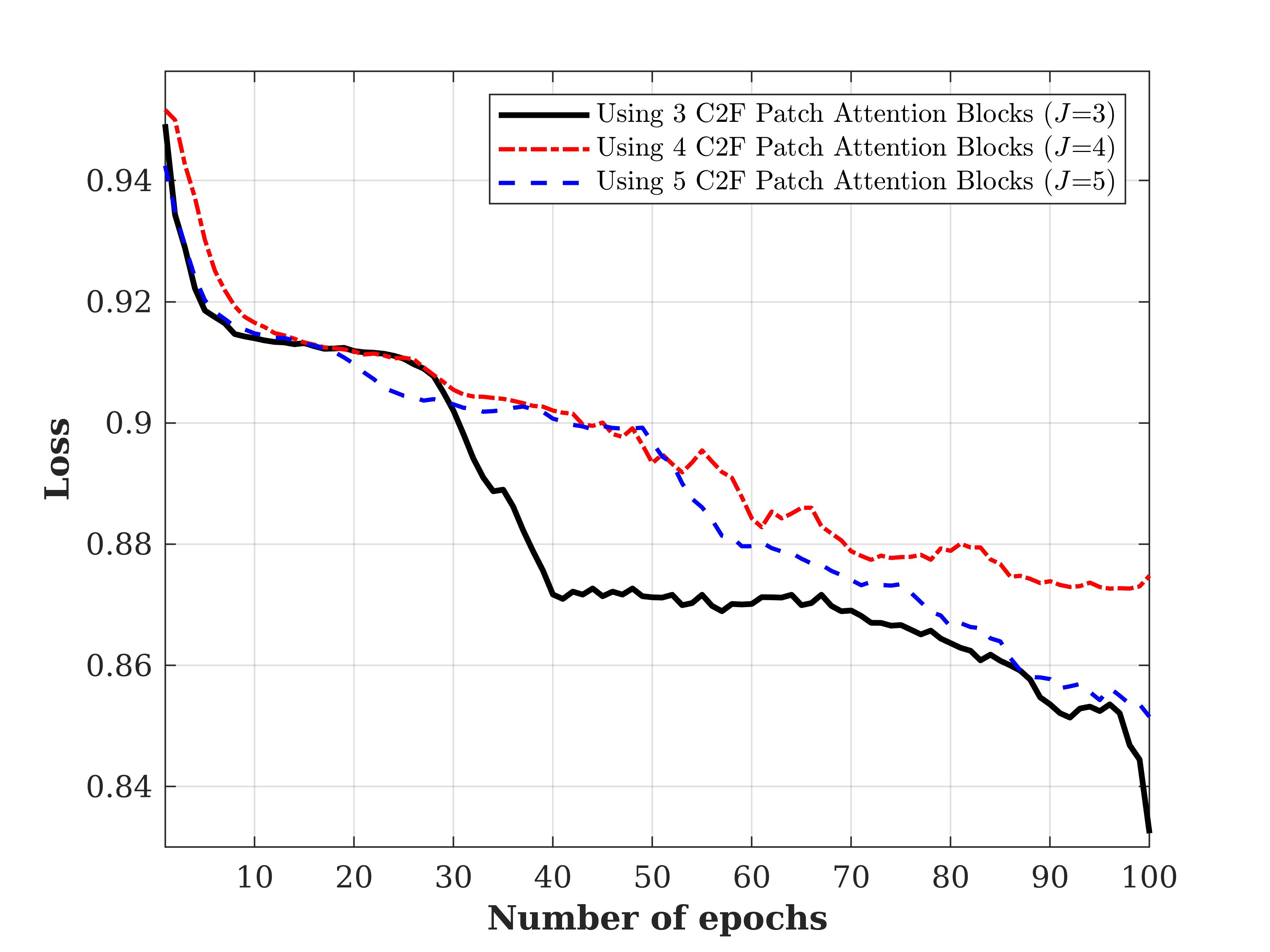}
\caption{Performance comparisons on different C2F patch attention blocks in the proposed network}
\label{sfig::sfig1}
\end{figure}

\clearpage
\subsection*{6.2 Multi-view 3D Reconstruction Results}
The performance comparisons of 3D-C2FT against benchmark models, namely 3D-R2N2 \cite{Choy2016}, Pix2Vox++/F \cite{Xie2020}, Pix2Vox++/A \cite{Xie2020}, and LegoFormer \cite{Yagubbayli2021} are presented in this subsection. As an illustration, we demonstrate several reconstructed instances from the ShapeNet dataset as shown in Supplementary Fig. \ref{sfig::sfig2}. The figures show that the objects reconstructed by the 3D-C2FT have complete and fine surfaces. In addition, we also demonstrate several results that were reconstructed by 3D-C2FT with different orientation views of refined 3D volumes in Video01.mp4.

\begin{figure}
\centering
\includegraphics[width=0.85\textwidth]{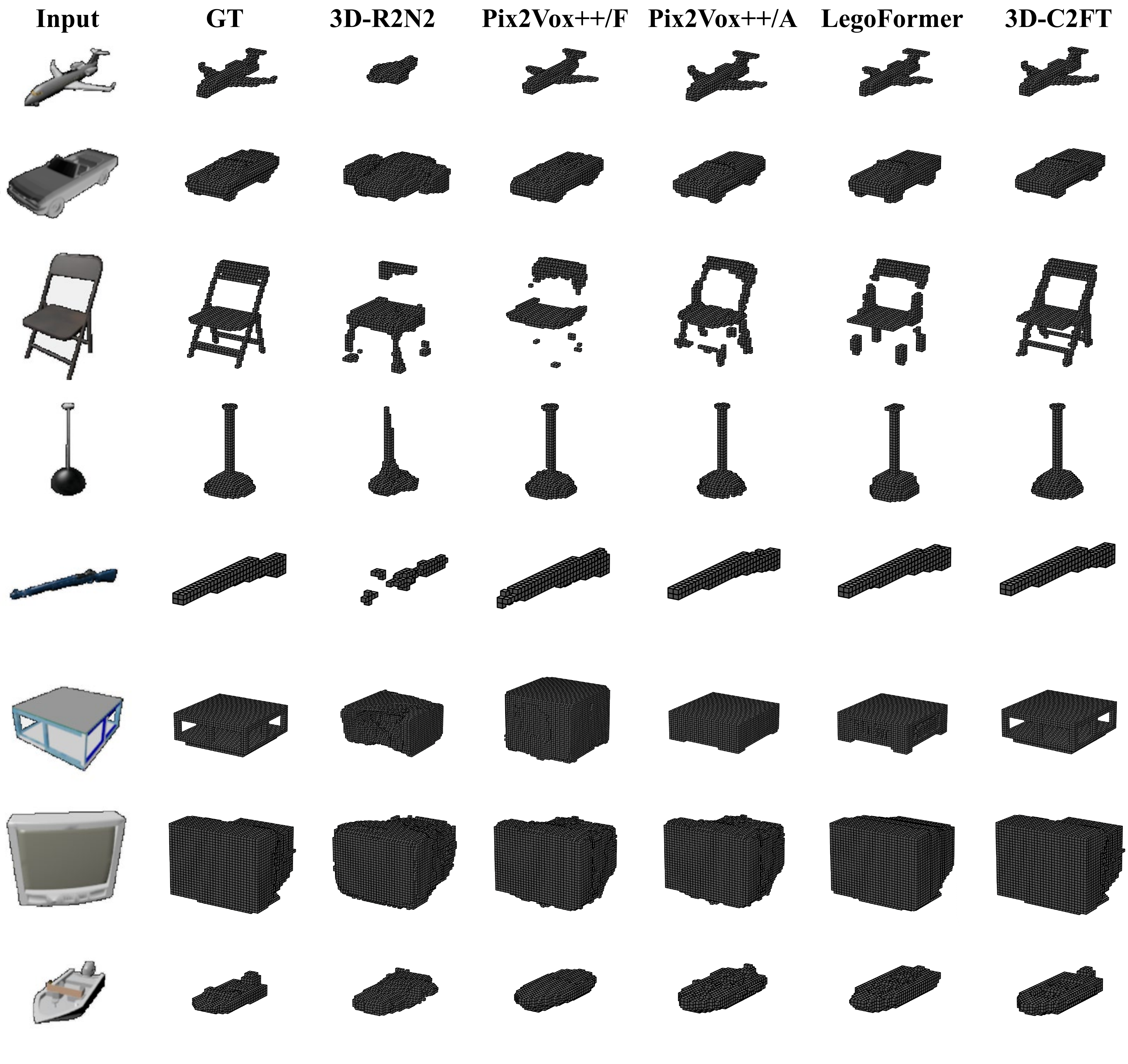}
\caption{3D object reconstruction on the ShapeNet dataset based on 12-view 2D images. GT is the ground truth of 3D volume}
\label{sfig::sfig2}
\end{figure}

\clearpage
\subsection*{6.3 Performance Analysis on Individual Categories}
We report the 3D reconstruction results on the ShapeNet dataset \cite{Wu2015} for individual categories and overall average. This experiment fixes the input views number at 12 for all models. Supplementary Table \ref{stab::stab2} reveals the proposed model attains the highest IoU value and F-score for most of the categories, except \textit{chair}, \textit{table}, \textit{telephone}, and \textit{watercraft}.

\begin{table}
\renewcommand{\arraystretch}{1.1}
\centering
\caption{Performance comparisons of 12-view 3D reconstruction results for each category on the ShapeNet dataset. The best score for each category is written in bold}
\label{stab::stab2}
\fontsize{7.5}{8}\selectfont
\begin{tabular}{l|C{0.8cm}C{1cm}C{0.8cm}C{1cm}C{0.8cm}C{1cm}C{0.8cm}C{1cm}C{0.8cm}C{1cm}} 
\hline
\multicolumn{1}{c|}{\multirow{3}{*}{\textbf{Category}}} & \multicolumn{10}{c}{\textbf{Model}} \\
\cline{2-11}
 & \multicolumn{2}{c}{3D-R2N2} & \multicolumn{2}{c}{Pix2Vox++/F} & \multicolumn{2}{c}{Pix2Vox++/A} & \multicolumn{2}{c}{LegoFormer} & \multicolumn{2}{c}{3D-C2FT} \\
\cline{2-11}
 & \textit{IoU} & \textit{F-score} & \textit{IoU} & \textit{F-score} & \textit{IoU} & \textit{F-score} & \textit{IoU} & \textit{F-score} & \textit{IoU} & \textit{F-score} \\ 
\hline
Airplane & 0.515 & 0.338 & 0.626 & 0.469 & 0.725 & 0.610 & 0.693 & 0.561 & \textbf{0.735} & \textbf{0.623} \\
Bench & 0.427 & 0.251 & 0.582 & 0.388 & 0.680 & 0.516 & 0.679 & 0.518 & \textbf{0.700} & \textbf{0.538} \\
Cabinet & 0.727 & 0.319 & 0.793 & 0.388 & 0.830 & 0.458 & 0.811 & 0.439 & \textbf{0.836} & \textbf{0.468} \\
Car & 0.806 & 0.440 & 0.850 & 0.516 & 0.882 & 0.595 & 0.877 & 0.576 & \textbf{0.883} & \textbf{0.598} \\
Chair & 0.483 & 0.190 & 0.591 & 0.263 & \textbf{0.644} & \textbf{0.338} & 0.629 & 0.320 & 0.641 & \textbf{0.338} \\
Display & 0.503 & 0.222 & 0.589 & 0.285 & 0.611 & 0.331 & 0.624 & 0.350 & \textbf{0.633} & \textbf{0.360} \\
Lamp & 0.375 & 0.229 & 0.477 & 0.314 & 0.493 & 0.352 & 0.489 & 0.335 & \textbf{0.516} & \textbf{0.360} \\
Loudspeaker \:& 0.683 & 0.239 & 0.736 & 0.277 & 0.760 & 0.323 & 0.751 & 0.317 & \textbf{0.771} & \textbf{0.336} \\
Rifle & 0.541 & 0.449 & 0.621 & 0.547 & 0.682 & 0.619 & 0.689 & 0.623 & \textbf{0.695} & \textbf{0.633} \\
Sofa & 0.641 & 0.290 & 0.714 & 0.336 & 0.779 & 0.448 & 0.779 & 0.447 & \textbf{0.782} & \textbf{0.452} \\
Table & 0.494 & 0.252 & 0.621 & 0.350 & \textbf{0.664} & \textbf{0.418} & 0.637 & 0.402 & 0.654 & 0.405 \\ 
Telephone & 0.669 & 0.409 & 0.803 & 0.583 & 0.846 & 0.662 & \textbf{0.854} & \textbf{0.676} & 0.842 & 0.650 \\ 
Watercraft & 0.557 & 0.301 & 0.616 & 0.366 & 0.666 & 0.458 & \textbf{0.680} & \textbf{0.479} & 0.671 & 0.466 \\ \hline
\textbf{Overall} & 0.636 & 0.382 & 0.692 & 0.449 & 0.717 & 0.460 & 0.717 & 0.470 & \textbf{0.720} & \textbf{0.474} \\
\hline
\end{tabular}
\end{table}

\clearpage
\subsection*{6.4 Occlusion Box}
\begin{figure}
\centering
\includegraphics[width=0.92\textwidth]{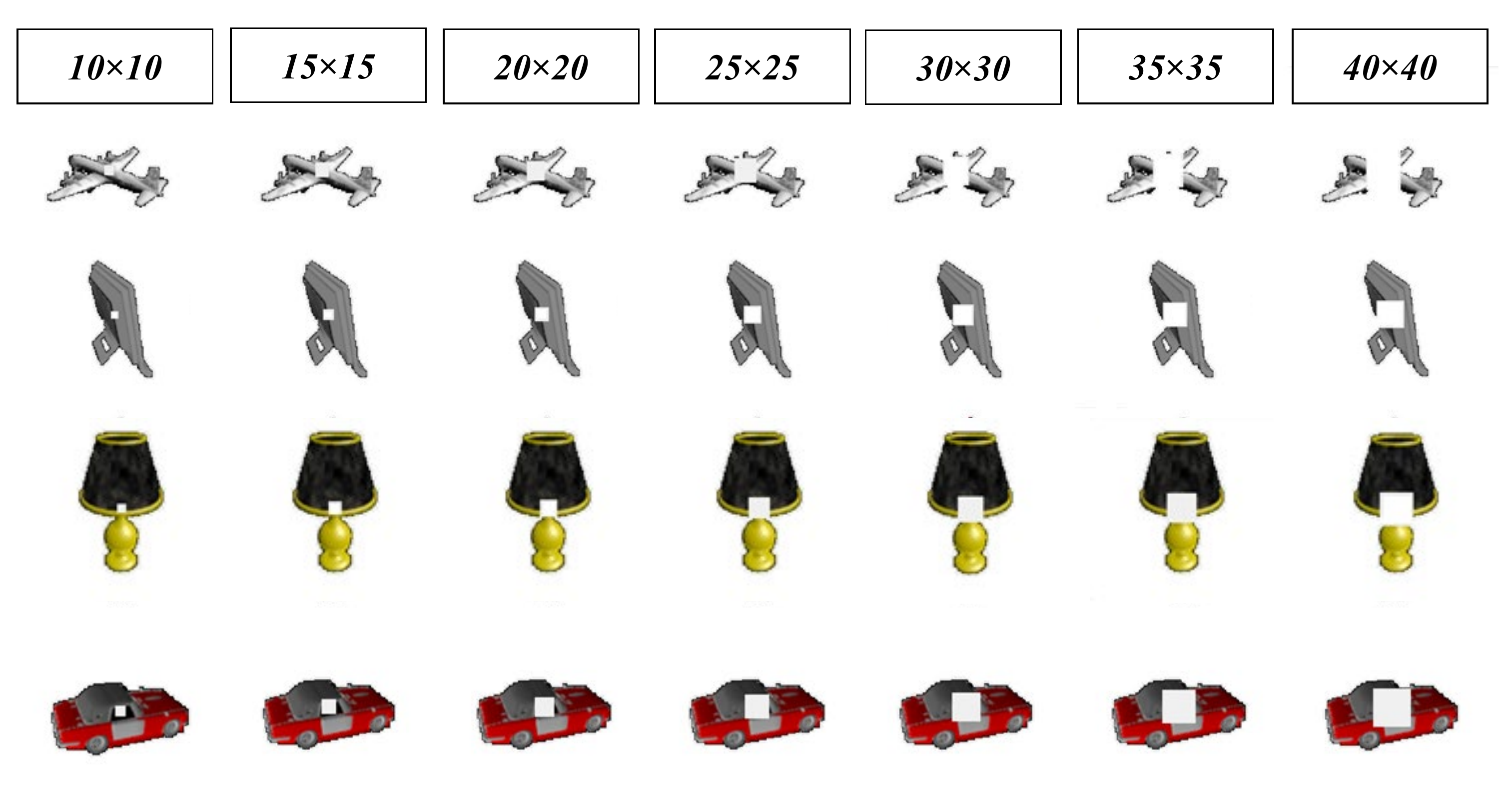}
\caption{Demonstrations of occlusion images on ShapeNet with several sizes of occlusion box: \textit{10}$\times$\textit{10}, \textit{15}$\times$\textit{15}, \textit{20}$\times$\textit{20}, \textit{25}$\times$\textit{25}, \textit{30}$\times$\textit{30}, \textit{35}$\times$\textit{35} and \textit{40}$\times$\textit{40}}
\label{sfig::sfig3}
\end{figure}

In this study, we introduce the occlusion box to the odd number ordered lists of 2D images, which impedes the important parts of the images. Supplementary Fig.\ref{sfig::sfig3} illustrates several test images from the ShapeNet dataset \cite{Wu2015} with different sizes of occlusion boxes.  

As an illustration, we demonstrate several reconstruction results on different sizes of occlusion boxes, as shown in Supplementary Fig. \ref{sfig::sfig4} and \ref{sfig::sfig5}. Among the CNN-based models and LegoFormer, 3D-C2FT performs the best over various occlusion boxes.

\begin{figure}
\centering
\includegraphics[width=0.90\textwidth]{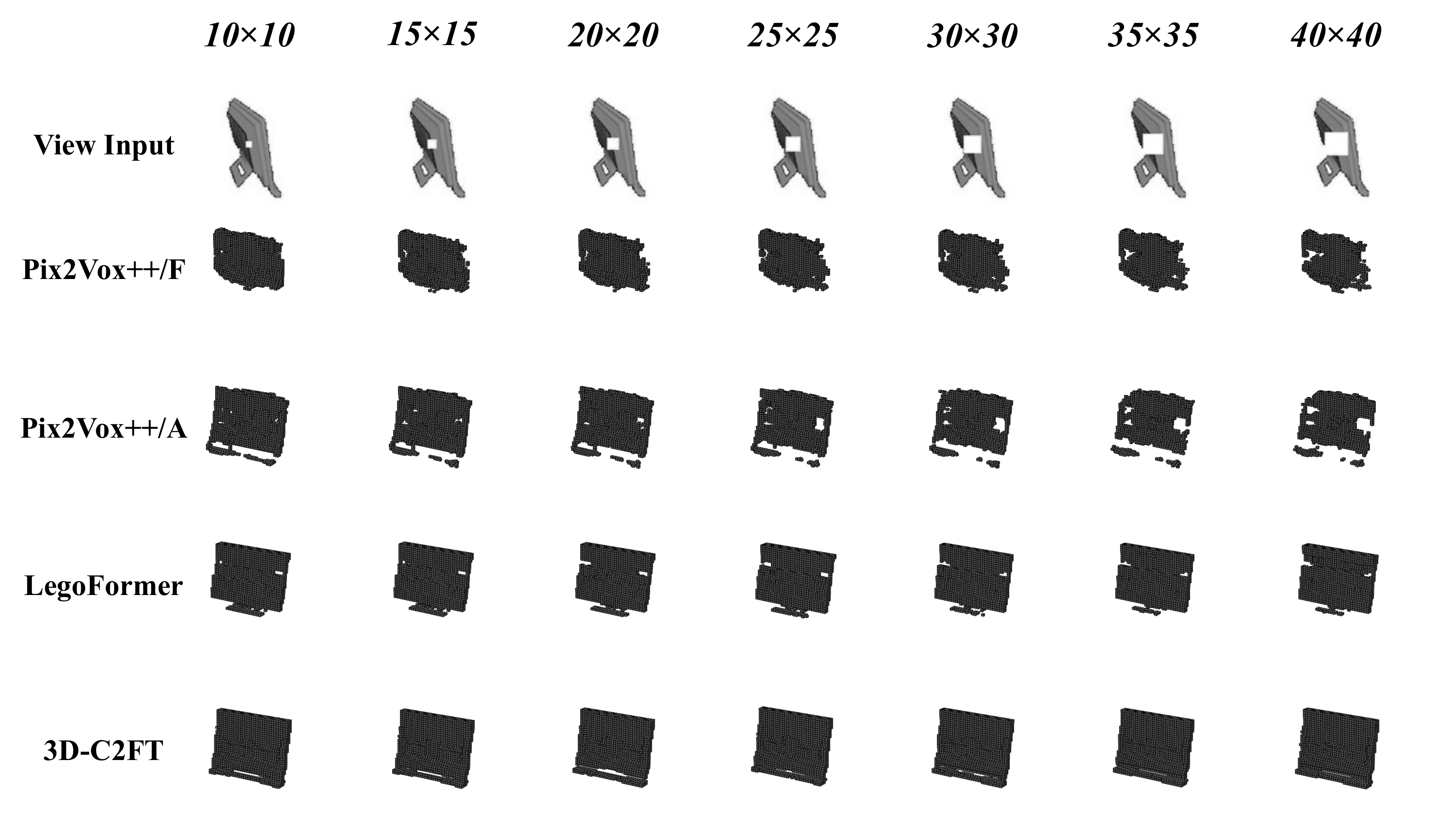}
\caption{Demonstration of 3D object reconstruction results for \textit{display} category with several sizes of occlusion box: \textit{10}$\times$\textit{10}, \textit{15}$\times$\textit{15}, \textit{20}$\times$\textit{20}, \textit{25}$\times$\textit{25}, \textit{30}$\times$\textit{30}, \textit{35}$\times$\textit{35} and \textit{40}$\times$\textit{40}. The experiments were conducted using 12 views (only 1 is shown)}
\label{sfig::sfig4}
\end{figure}

\begin{figure}
\centering
\includegraphics[width=0.90\textwidth]{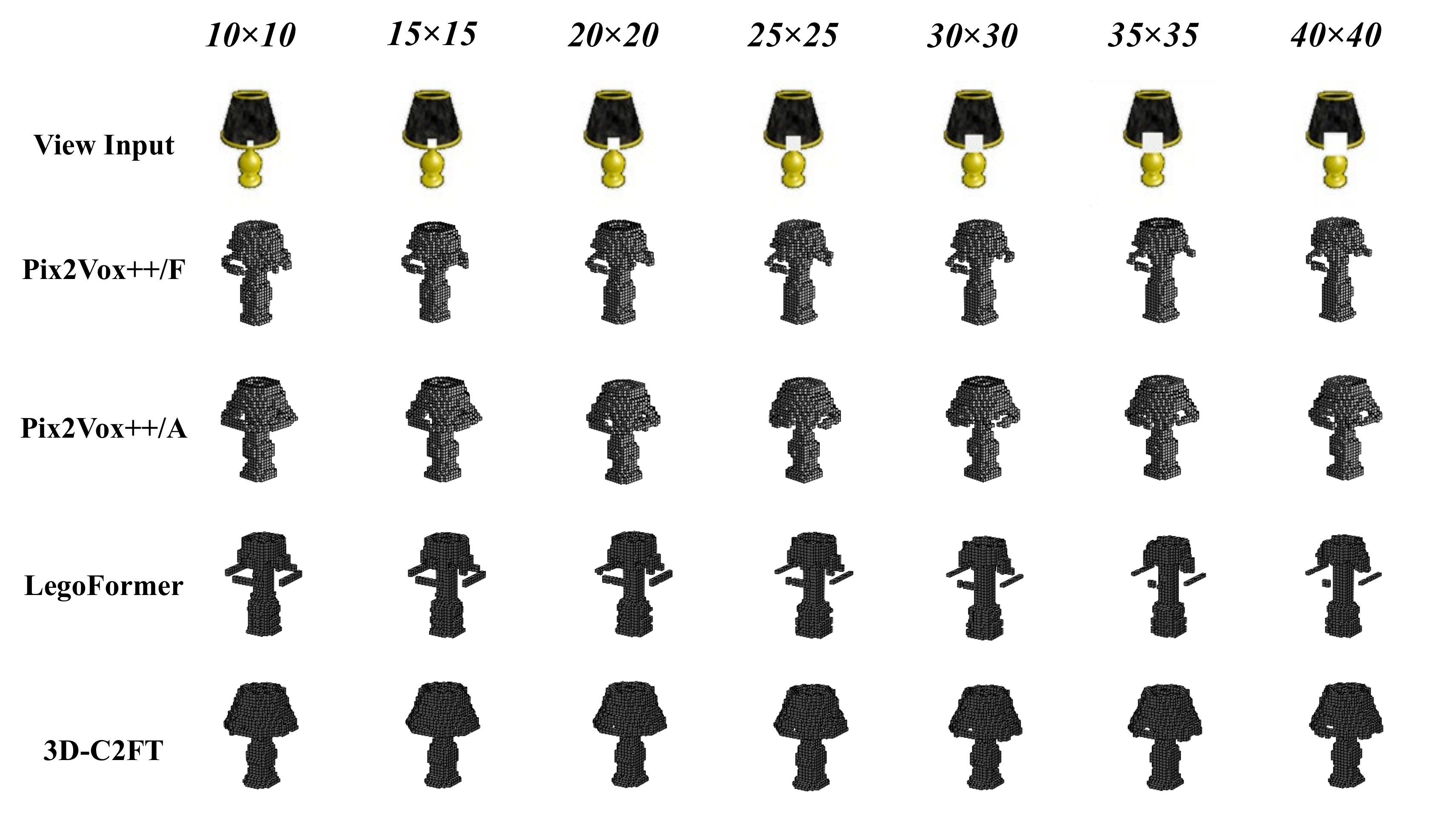}
\caption{Demonstration of 3D object reconstruction results for \textit{lamp} category with several sizes of occlusion box. The experiments were conducted using 12 views (only one is shown)}
\label{sfig::sfig5}
\end{figure}

\clearpage
\subsection*{6.5 Experiments on Multi-view Real-life Dataset}
This section conducts additional qualitative experiments for single and multi-view 3D reconstruction with a Multi-view Real-life dataset. 

\begin{figure}
\centering
\includegraphics[width=0.85\textwidth]{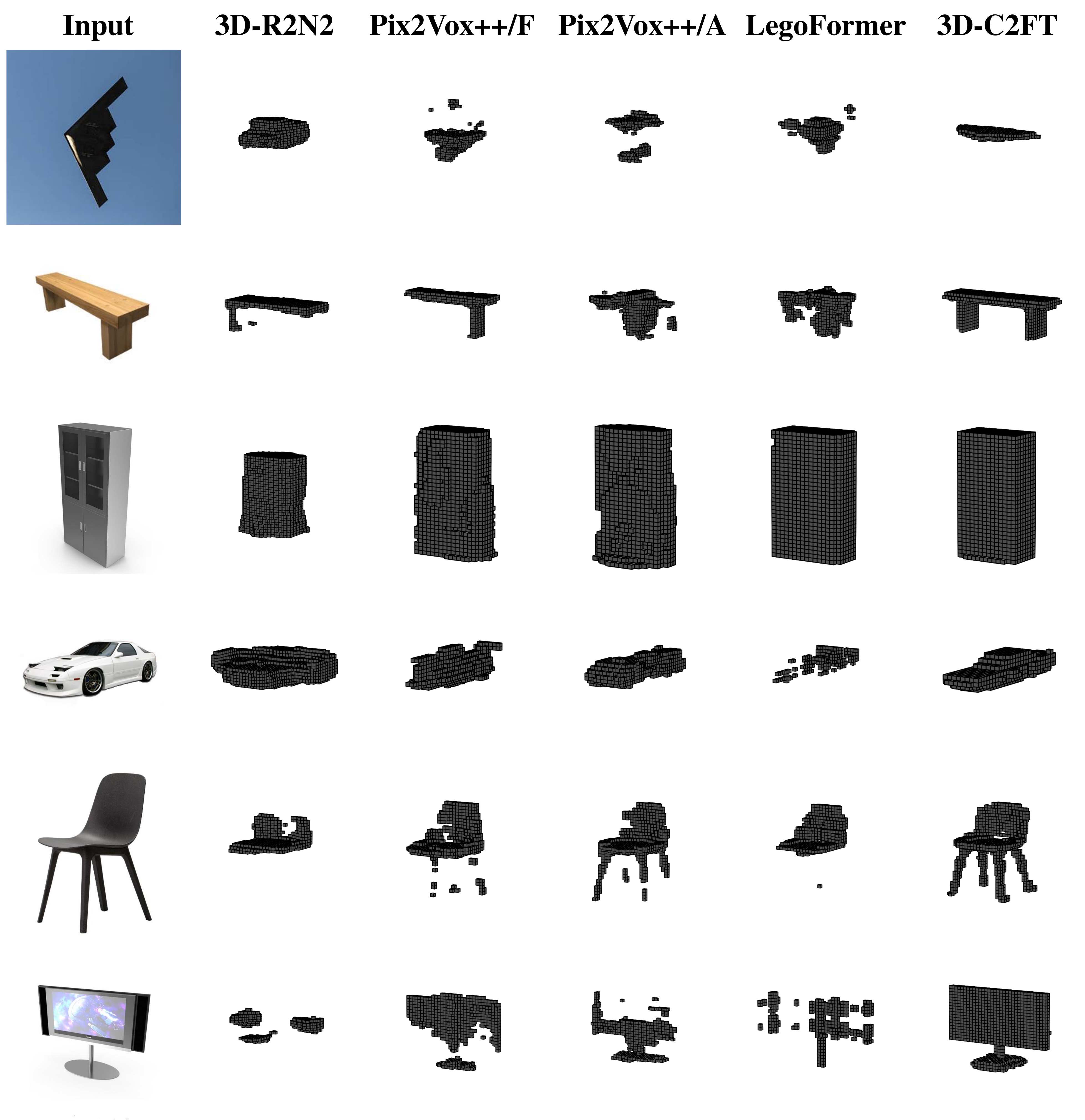}
\caption{3D object reconstruction results for \textit{Case I}. Each test sample mainly contains 1 to 5 views}
\label{sfig::sfig6}
\end{figure}

Supplementary Fig.\ref{sfig::sfig6} demonstrates the performance comparisons on \textit{Case I}. Our proposed model performs substantially better than the benchmark models in terms of the refined 3D volume and surface quality. In particular, LegoFormer performs very poorly for all categories, except \textit{cabinet}, which is comparable to the 3D-C2FT.

\begin{figure}
\centering
\includegraphics[width=0.85\textwidth]{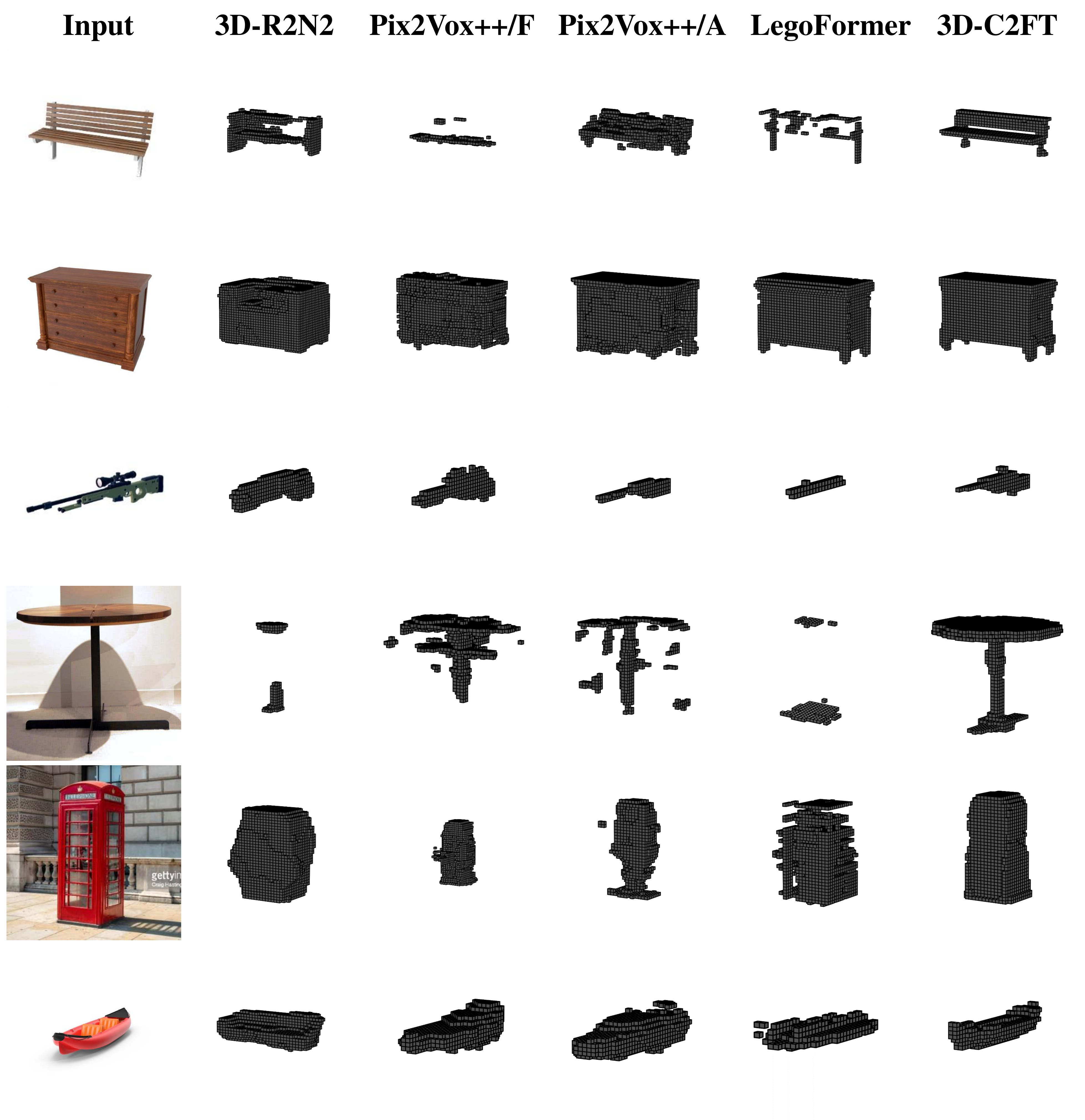}
\caption{3D object reconstruction results for \textit{Case II}. Each test sample mainly contains at least 6 to 11 views}
\label{sfig::sfig7}
\end{figure}

\clearpage
In addition, Supplementary Fig. \ref{sfig::sfig7} illustrates the performance comparisons in \textit{Case II}. 3D-C2FT remains significantly better than the benchmark models. Pix2Vox++/A and LegoFormer were slightly improved in this case compared to Case I. Although both models can barely reconstruct the 3D objects, the fineness is not on par with 3D-C2FT.

\begin{figure}[!t]
\centering
\includegraphics[width=0.85\textwidth]{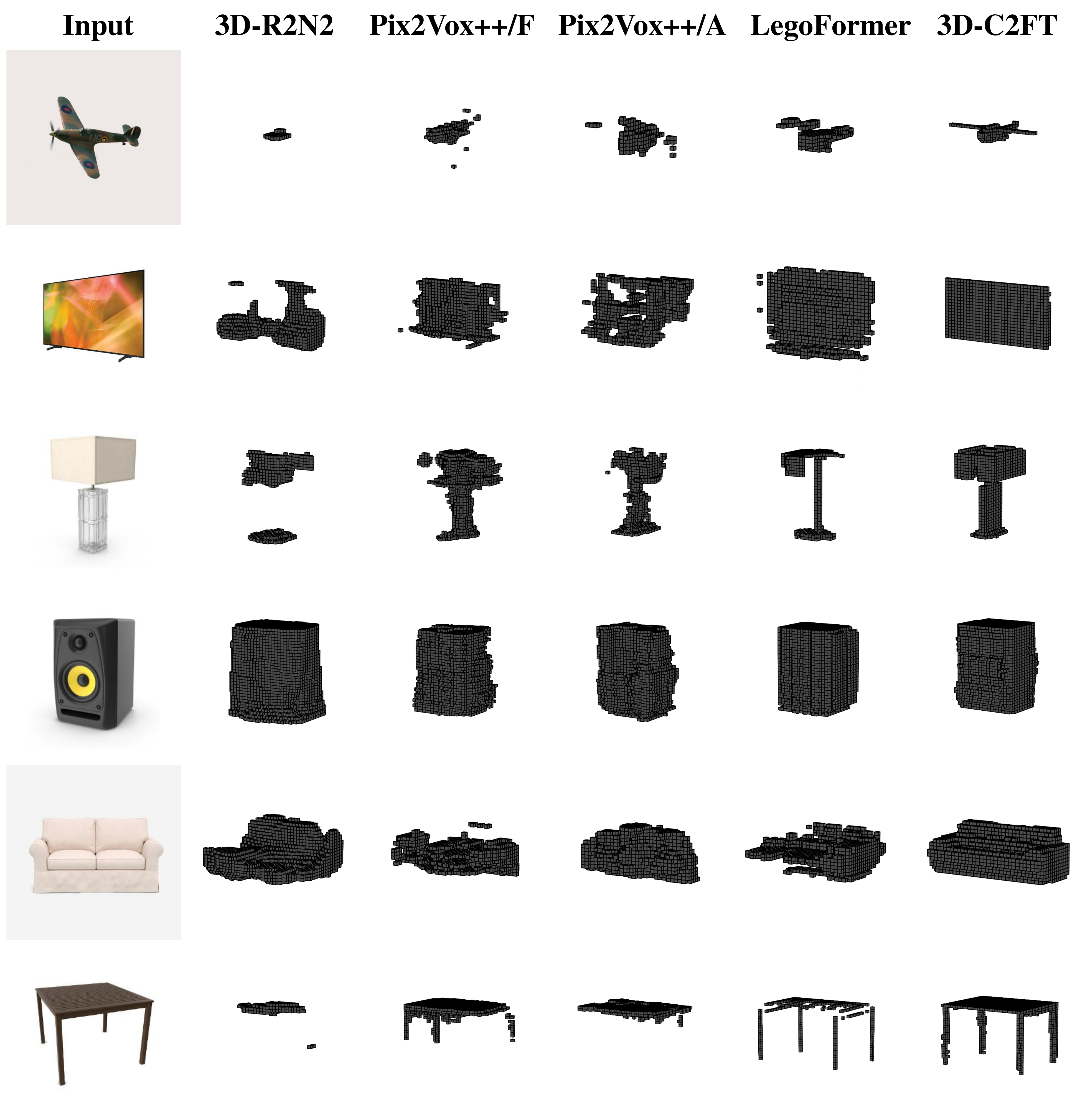}
\caption{3D object reconstruction results for \textit{Case III}. Each test sample mainly contains at least 12 views or more}
\label{sfig::sfig8}
\end{figure}

For \textit{Case III}, Supplementary Fig.\ref{sfig::sfig8} illustrates several qualitative examples of 3D reconstruction by using more than 12 views. The reconstructed objects by 3D-C2FT are more well structured with all fine-grained details. However, surprisingly, most competing models perform poorly under \textit{Case III}. For instance, the 3D volumes are not appropriately reconstructed for most categories, or the surfaces of 3D objects are not generated correctly.

In summary, from the analysis of Supplementary Fig. \ref{sfig::sfig6}--\ref{sfig::sfig8}, we find all the competing models cannot reconstruct the real-life 3D objects decently. Notably, the proposed model performs pretty well even in \textit{Case I}, which is considered the most challenging one. This observation again vindicates the effectiveness of the C2F attention mechanism applied in the 3D-C2FT. We additionally demonstrate several sample results in Video02.mp4.

\end{document}